\documentclass{article}
\usepackage[a4paper,margin=2cm]{geometry}

\usepackage[utf8]{inputenc}
\usepackage{lineno}
\usepackage{graphicx}
\usepackage{amsmath}
\usepackage{amsfonts}
\usepackage{amssymb}
\usepackage{amsthm}
\usepackage{subfig}
\usepackage{natbib}
\usepackage{todonotes}
\usepackage{url}
\usepackage[boxed,linesnumbered]{algorithm2e}
\usepackage{setspace} 
\usepackage{array}
\usepackage{comment}
\usepackage{soul}
\usepackage{yhmath}
\usepackage{enumerate}
\usepackage{morefloats}
\usepackage{mathtools}
\usepackage{hyperref}

\usepackage{dsfont}

\DeclareMathOperator{\atantwo}{atan2}

\begin{document}
  
  \title{Congestion control algorithms for robotic swarms with a common target based on the throughput of the target area}

  \author{
    Yuri Tavares dos Passos\footnote{Corresponding author}
    \\
    yuri.passos@ufrb.edu.br
    \and
    Xavier Duquesne
    \\ duquesne.xavier.13@gmail.com 
    \and
    Leandro Soriano Marcolino
    \\ l.marcolino@lancaster.ac.uk
  }

  \maketitle
  
  \begin{abstract}
    When a large number of robots try to reach a common area, congestions happen, causing severe delays. To minimise congestion in a robotic swarm system, traffic control algorithms must be employed in a decentralised manner. Based on strategies aimed to maximise the throughput of the common target area, we developed two novel algorithms for robots using artificial potential fields for obstacle avoidance and navigation. One algorithm is inspired by creating a queue to get to the target area (Single Queue Former -- SQF), while the other makes the robots touch the boundary of the circular area by using vector fields (Touch and Run Vector Fields -- TRVF). We performed simulation experiments to show that the proposed algorithms are bounded by the throughput of their inspired theoretical strategies and compare the two novel algorithms with state-of-art algorithms for the same problem (PCC, EE and PCC-EE). The SQF algorithm significantly outperforms all other algorithms for a large number of robots or when the circular target region radius is small. TRVF, on the other hand, is better than SQF only for a limited number of robots and outperforms only PCC for numerous robots. However, it allows us to analyse the potential impacts on the throughput when transferring an idea from a theoretical strategy to a concrete algorithm that considers changing linear speeds and distances between robots.
  \end{abstract}

\section{Introduction}

To be more cost-effective, an increasing number of problems are tackled using many simple robots rather than a few complex robots \citep{583135}. Systems formed by a large number of robots are called robotic swarms. They are composed of robots that can only interact with direct neighbours and follow simple rules \citep{10.1007/978-3-642-58069-7_38}. Counter-intuitively, some combinations of simple rules can create complex global behaviours \citep{DBLP:journals/adb/Mataric95,Liviu2003}. Hence, some of the robotic swarm research is based on the creation of complex behaviours based on simple rules \citep{navarro2013introduction,Garnier2007}. Therefore, careful design can enable swarms of simple robots to perform complex tasks with robustness instead of using complex and expensive robots.

However, robotic swarms are often slowed down by conflicts created by the trajectories of the robots. For example, traffic congestion appears when several robots try to reach the same area simultaneously \citep{10.1145/1141911.1142008, 6094640, survey107, grossman}. Congestion happens when several robots need to reach a common area,
such as in waypoint navigation \citep{marcolinoNoRobotLeft2008,10.1371/journal.pone.0151834} and in foraging \citep{ducatelleCommunicationAssistedNavigation2011,Fujisawa2019}.
In the current state of the art, congestion problems in robotic swarms are mainly solved by collision avoidance in a decentralised manner because it enables better scalability of the algorithms \cite{pmlr-v164-batra22a,8594422,BORRMANN201568}.
Despite that, just avoiding collisions does not necessarily lead to a good performance in the common target problem. For example, \citet{Marcolino2016} showed that using only the decentralised collision avoidance algorithm ORCA (Optimal Reciprocal Collision Avoidance)  
\citep{Berg2011} creates local minima around the common target region. They present experimental solutions for the congestion problem that arises when the swarm shares the same goal, but we found that these algorithms may completely fail in small target regions, with an area measuring less than five times the area of the robots. Moreover, evolution-based algorithms may adapt to congestions after thousands of iterations \cite{Fujisawa2019}, but we focus on algorithms directly designed to handle congestion without requiring long iterations for training or adaptation.

Additionally, as congestion in robotic swarms grows with the number of individuals, any analysis of the efficiency of algorithms to reduce congestion needs to incorporate the increase in the number of robots. In this regard, we introduced in our concurrent work \citep{arxivTheory} a viable metric for analysing large-scale swarm robotics systems: the throughput of the common target area. We defined it as the inverse of the average time between arrivals at the target area. Furthermore, in that work, we have developed theoretical strategies to maximise the throughput in scenarios where the robots have constant maximum linear speed and fixed minimum distance between each other. Although these scenarios are useful for ease of mathematical analysis, they are not practical in real applications as inter-robot distances and speeds vary, but the presented strategies help design new algorithms. 

Therefore, the main contribution of this paper is that, based on these theoretical strategies, we present two novel algorithms for handling congestion in more realistic settings: Single Queue Former (SQF) and Touch and Run Vector Fields (TRVF).  We evaluate our algorithms by realistic Stage \citep{PlayerStage} simulations with holonomic and non-holonomic robots. From these simulations, we conclude that SQF outperforms the state-of-the-art and can handle a small target region. In that situation, previous approaches completely failed in our simulations. Moreover, we learn that TRVF helps us to understand the potential impacts of the variation in the linear speed and distance between the robots when translating an idea from theoretical strategies to concrete algorithms.

This paper is organised as follows. In the next section, we discuss some related work. We describe the two algorithms in Section \ref{sec:proposedalgs}. Section \ref{sec:experimentresults} presents the experiments and compares the two new algorithms with the state-of-art ones. Finally, we summarise our results and make final remarks in Section \ref{sec:conclusion}.

\section{Related Work}


Robotic traffic control has been studied for a long time \citep{grossman,kato92,CaloudIndoor}, but assuming robots navigating in delimited lanes and coordination is necessary only at the intersections. Other more recent works still focus on alleviating congestions in delimited lanes and circuits \citep{9208701,hoshinoMultirobotCoordinationJams2013,hoshinoMultiRobotCoordinationMethodology,viswanathLoadbalancedDecongestedMultirobotic2009}. For instance,  an algorithm based on Petri nets has been introduced to avoid deadlocks at intersections \citep{zhouCollisionDeadlockAvoidance2017}, where each robot follows a pre-determined closed path, but it works only for robots on lanes and does not show performance in terms of simulation time nor throughput. \citet{Saska2020} developed a control algorithm for motion planning of formations for unmanned aerial vehicles in environments with narrow passages. The robots in this system follow a leader and maintain proximity while avoiding obstacles in a constrained space. \citet{Yoshimoto2018} describe a decentralised algorithm to maintain proximity between robots. Although their work is designed for a robotic swarm, it is also leader-based. 

Similarly, there are many relevant works in the multi-agent systems literature, but assuming autonomous cars that navigate following lanes or roads, and coordination is needed at the junctions \citep{carlinoAuctionbasedAutonomousIntersection2013,Sharon2017}. Other lines of work try to optimise trajectories across edges of complex traffic networks, for example, by global planning and using incentives or tolls for self-interested agents \citep{Sharon2017b,Sharon2018}. \citet{7798529} use the Kuramoto model to coordinate multiple vehicles towards a target while avoiding collisions and keeping them next to each other, but they do not analyse the time to get to the target and exit from it as the number of vehicles grows. \citet{Ma2020} describe a reinforcement learning algorithm for multi-agent planning for a swarm of vehicles to go to objectives distributed over space, but it does not perform well when the targets cannot be well divided into regions over the environment.

The problem of alleviating congestion when a swarm of robots has a common target has not yet been well studied. Thorough surveys about swarm robotics \citep{sahin04swarm,SahinGBT08,Barca2013swarm,Brambilla2013Swarm,Bayindir2016,8424838} 
do not discuss these situations. Additionally, a recent survey on collision avoidance \citep{hoyAlgorithmsCollisionfreeNavigation2015} provides insights into multiple vehicle navigation, but they do not discuss this issue in multi-robot navigation.

Collision avoidance is also an important related topic of study in robotics. We can find in the literature algorithms that let robots move efficiently in environments with a lot of obstacles, but they do not measure the efficiency of the algorithm in the common target problem \citep{9789280}. As an example, \citet{ferreraDecentralizedSafeConflict2017} have proposed a decentralised algorithm where a set of local reactive rules are followed to avoid collisions. We also note that their work shows an extensive list of benchmarks and metrics for robotics problems, but performance in common target situations is not included. Evolution-based algorithms may learn traffic rules for collision avoidance after thousands of iterations \cite{Fujisawa2019}, but they were not yet explored for the common target problem. Additionally, in this paper, we focus on algorithms directly designed to handle congestion based on our theoretical work \cite{arxivTheory} without requiring long iterations of training or optimisation.

However, only avoiding collisions does not necessarily lead to a good performance in the common target problem. For instance, \citet{Marcolino2016} showed that the ORCA algorithm \citep{Berg2011} reaches an equilibrium where the robots cannot go to the target. They proposed three algorithms using artificial potential fields and offered experimental solutions for the target congestion problem. However, we found that their proposed algorithms face issues as the target area gets smaller. 

Inspired by theoretical developments in our concurrent work \citep{arxivTheory}, we propose new algorithms that outperform these previous approaches. That is, we proposed to analyse the arrival rate in the target area as the time tends to infinity as an alternative approach to analysing congestion in swarms with a common target area. This measure the advantages of being finite for any closed target region of any shape as the number of robots grows, as opposed to the number of robots per time of arrival at the target area. Assuming a circular target area and robots with constant maximum linear speed and fixed minimum distance from each other, we developed theoretical strategies for entering that area and calculated their theoretical throughput for a given time and their asymptotic throughput. The presented theoretical strategies were based on forming a corridor towards the target area or making multiple curved trajectories towards the boundary of the target area. However, they assumed constant speeds and fixed minimum distances between robots, thus not yet directly applicable in practice. Hence, based on the corridor strategy, we propose here the Single Queue Former algorithm, and, inspired by the multiple curved trajectories, we introduce the Touch and Run Vector Fields algorithm.

\section{Proposed algorithms}
\label{sec:proposedalgs}

In this section, we consider the scenario where a large number of robots must reach a common target. After reaching the target, each robot moves towards another destination which may or may not be common among the robots. We assume the target is defined by a circular area of radius $s$. A robot reaches the target if its centre of mass is at a distance below or equal to the radius $s$ from the centre of the target. We assume that there is no minimum amount of time to stay at the target. 

We present two algorithms based on the results in \citep{arxivTheory}. In it, the \emph{throughput} of the target area is used to measure performance, which is defined as the inverse of the average time between arrivals at the target. Differently to that work, which intended to find the maximum throughput for the strategies by using constant maximum linear speed and fixed minimum distance between the robots, in real applications, robots move at variable speeds and do not always keep the same distance between others, although they may try to maintain some distance to avoid bumping. In addition to that, the space around the target region is finite and may have other obstacles.

Having this in mind, we devised two algorithms: the Single Queue Former and the Touch and Run Vector Fields. These algorithms are only applied inside a circle of radius $D > s$ around the target. Also, we assume that if the robots have two or more targets, they are apart from each other at least $2D$.

We hereafter use geometric notation as follows. $\overrightarrow{AB}$ and $\overline{AB}$ represents a ray starting at A and passing through B and a segment from A to B, respectively. $|\overline{AB}|$ is the size of $\overline{AB}$. If a two-dimensional point is represented by a vector $P_{1}$, its x- and y-coordinates are denoted by $P_{1,x}$ and $P_{1,y}$, respectively. $\bigtriangleup ABC$ expresses the triangle formed by the points A, B and C. $\widehat{AOB}$ means an angle with vertex O, one ray passing through point A and another through B. All angles are measured in radians in this paper. 

\subsection{Single Queue Former}

Based on the corridor strategy from \citep{arxivTheory}, we propose an algorithm, named Single Queue Former (SQF), to improve throughput in case of target congestion whose aim is to form a single queue that goes towards the target. The queue is formed inside a rectangular corridor of width equal to the circular target diameter and a given fixed length. The robots should only enter the target region by this queue. 

Specifically, this algorithm is based on the compact lanes and hexagonal packing strategies findings from \citep{arxivTheory}. However, we do not enforce the hexagonal formation here to avoid robots losing time forming it, although the hexagon packing is the configuration of the robots that maximises the throughput in an ideal scenario where the robots maintain the same distance and keep constant linear speed \citep{arxivTheory}. As it will be shown in Section \ref{sec:experimentresults}, the throughput for this algorithm is bounded by the hexagonal packing throughput for the mean distance and mean linear speed achieved by the robots. 

\begin{figure}[t]
  \centering
  \includegraphics[width=0.45\columnwidth]{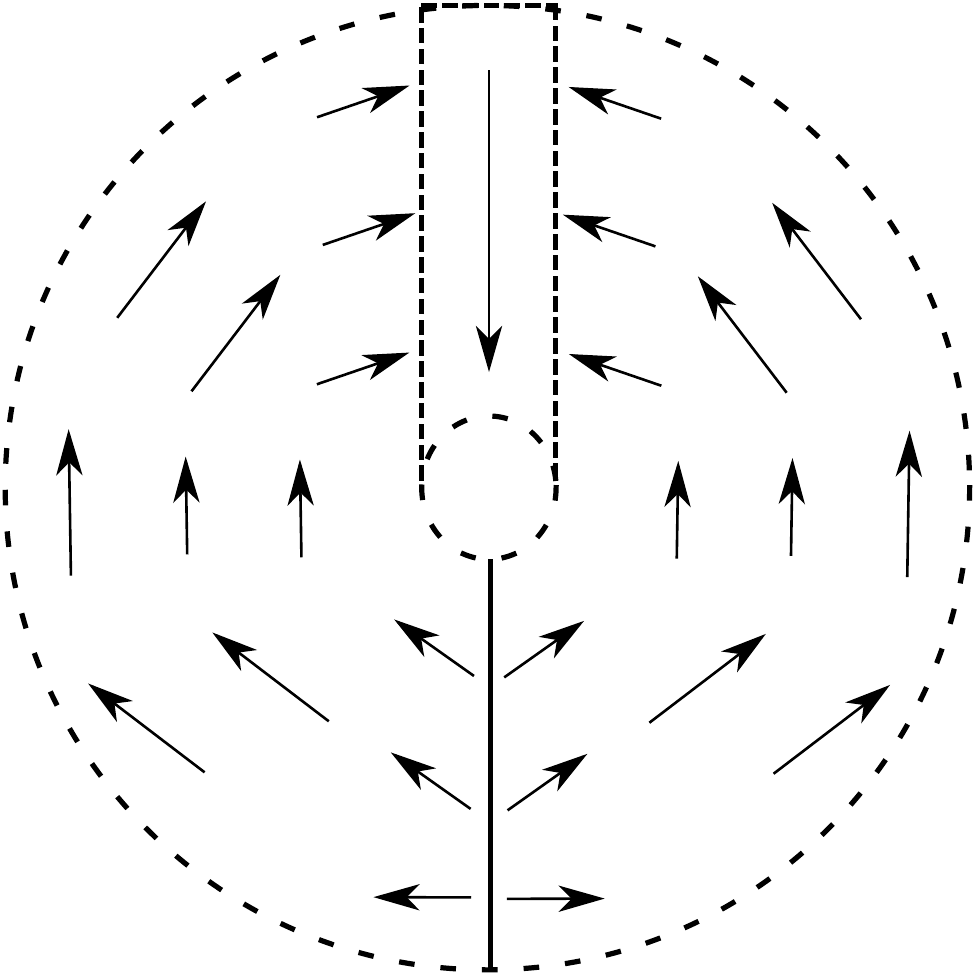}
  \caption{Rotational force field to reach the target with Single Queue Former algorithm.}
  \label{fig:sqf:entry_path}
\end{figure}

We use artificial potential fields \citep{Siegwart2004Introduction} to apply forces on the robots to form the queue and exit the target area efficiently. Let $s$ be the radius of the current target.
The corridor queue has a width $2s$ and starts from the current target centre, denoted by $o$.
Let the length of the corridor be the same value used for the radius of the circular working area of the algorithms around the target, denoted $D$.
Without any loss of generality, the corridor is located between the point $(o_x, o_y)$
and the point $t = (o_x, o_y + D)$.
Robots that are going to enter the target area are submitted to a rotational field whose centre is the position of the target centre.
When they reach the corridor, they are submitted to an attractive force towards the target. An illustration is shown in Figure \ref{fig:sqf:entry_path}.

Therefore, robots outside the corridor are submitted to a force according to the following equations. 
For the right-hand side of the circle, an anti-clockwise rotational field:

\begin{equation}
F_{x}  = - K_{SQF} \frac{p_y - o_y}{\Vert p - o \Vert},
\quad
F_{y}  = + K_{SQF} \frac{p_x - o_x}{\Vert p - o \Vert}.
\label{eq:vortex11}
\end{equation}
For the left-hand side of the circle, a clockwise rotational field:

\begin{equation}
F_{x}  = + K_{SQF} \frac{p_y - o_y}{\Vert p - o \Vert},
\quad 
F_{y}  = - K_{SQF} \frac{p_x - o_x}{\Vert p - o \Vert},
\label{eq:vortex12}
\end{equation}
where $F_{x}$ and $F_{y}$ are the two components of the force applied, 
$p = (p_x, p_y)$ is the position of the robot,
$o = (o_{x},o_{y})$ is the target centre, and 
$K_{SQF}$ is a constant for setting the force magnitude.

\begin{figure}[t]
  \centering
  \includegraphics[width=0.45\columnwidth]{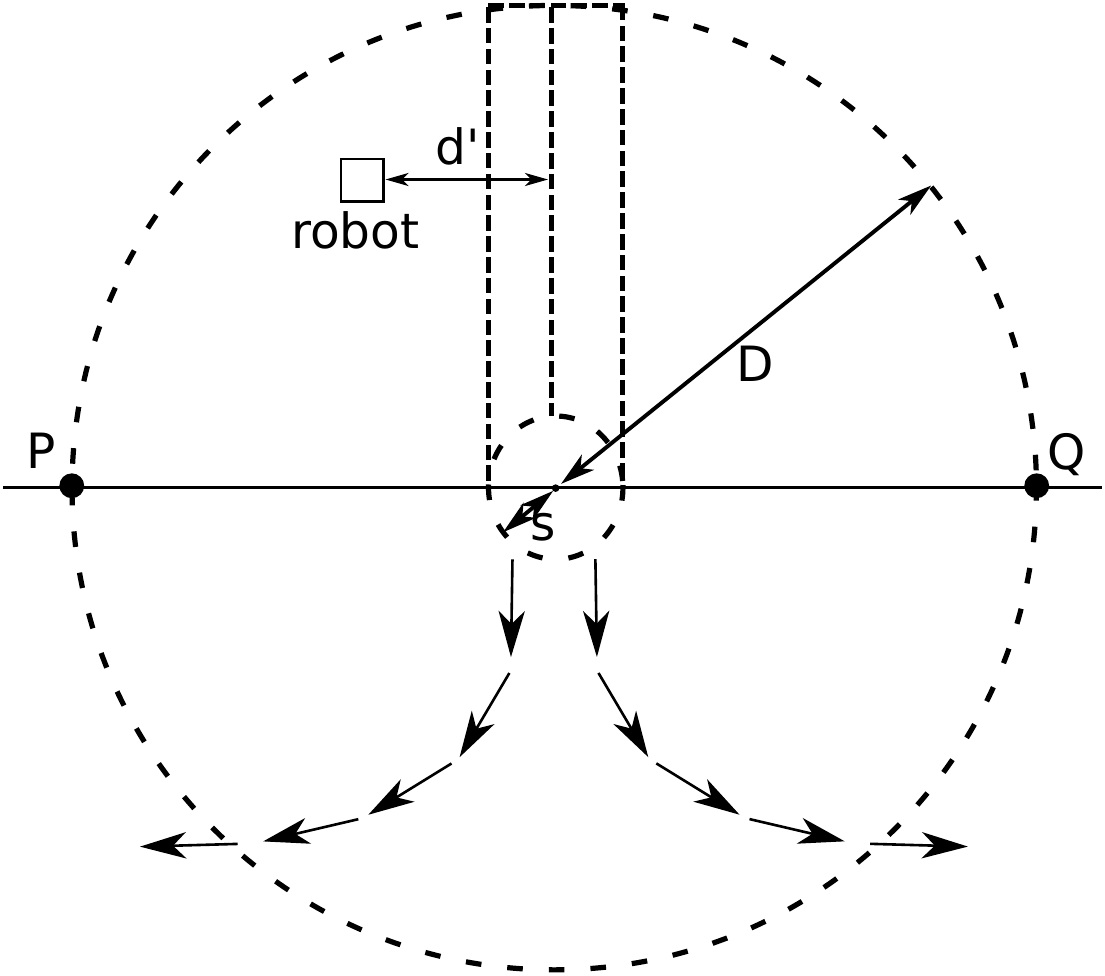}
  \caption{Rotational force field to leave the target in the SQF algorithm, and the distance $d'$ from the central line of the corridor to the robot for varying its influence radius.}
  \label{fig:sqf:exit_path}
\end{figure}

Additionally, we noticed in previous algorithms \citep{Marcolino2016} that
the robots reaching the target tend to stop on the target and orient themselves towards their next goal. 
This increases the time during which the target is occupied and thus decreases the throughput.
To solve this problem,  robots exiting nearby the target are constrained by another rotational field. This field also is only applied inside a circle with radius $D$ around the target (Figure \ref{fig:sqf:exit_path}). 
The aim of this field is to be aligned with the corridor orientation at the target and progressively allow robots to change directions. This field also depends on the new target position, not only on the robot's location, because if some robots reach the target area on a side but have to go to the opposite side when they leave the circle, it will cause congestion next to the exit area. For a robot with a new target located on the right-hand side of the previous target centre, we apply an anti-clockwise rotational field:

\begin{equation}
F_{x}  = - K_{SQF} \frac{p_y - Q_y}{\Vert p - Q \Vert},
\quad 
F_{y}  = + K_{SQF} \frac{p_x - Q_x}{\Vert p - Q \Vert}.
\label{eq:vortex21}
\end{equation}
For a robot with a new target located on the left-hand side of the previous target centre, a clockwise rotational field is applied:

\begin{equation}
F_{x}  = + K_{SQF} \frac{p_y - P_y}{\Vert p - P \Vert},
\quad
F_{y}  = - K_{SQF} \frac{p_x - P_x}{\Vert p - P \Vert},
\label{eq:vortex22}
\end{equation}
where $Q = (Q_x, Q_y) = (o_x + D, o_y)$ is a point on the right of the target, and $P = (P_x, P_y) = (o_x - D, o_y)$ is a point on the left of the target.

This outgoing rotational field enables robots to stay on their course after they reach the target 
and gradually rotate towards their next goal.
The orientation of the field depends on the position of the robot relative to the corridor. 
If the robot is on the left of the corridor,  the robot follows the rotational field going to the left with centre $P$. Otherwise, 
if it is on the right of the corridor, it follows the rotational field going to the right with centre $Q$.

Finally, let the influence radius be the maximum distance a robot considers anything sensed as an obstacle to avoid from its mass centre. To enable the robots to deal with small target sizes, we use two different constants for the influence radius between the robots when calculating repulsive forces, $I_{d}$ and $I_{min}$, with $I_{min} < I_{d}$: $I_{d}$ is the default and maximum radius for influence, and $I_{min}$ is the minimum allowed.

For robots inside the corridor or exiting the target region, the influence radius is set to $I_{min}$. Now, consider a robot is outside the corridor, but above the $\overline{PQ}$ line in Figure \ref{fig:sqf:exit_path}. Let $d'$ be the distance between the robot and the central line of the corridor (that is, from the position of the robot to the closest point in the vertical axis of Figure \ref{fig:sqf:exit_path}, starting from the target centre). Its influence radius $I$ varies in relation to $d'$ and is set to $I = I_{min} + d'$ only for $0 \le d' \le I_{d} - I_{min}$. This range for $d'$ guarantees that $I_{min} \le I \le I_{d}$. For the rest of the robots the influence radius is $I_{d}$.

\begin{figure}[h]
  \centering
  \includegraphics[width=\columnwidth]{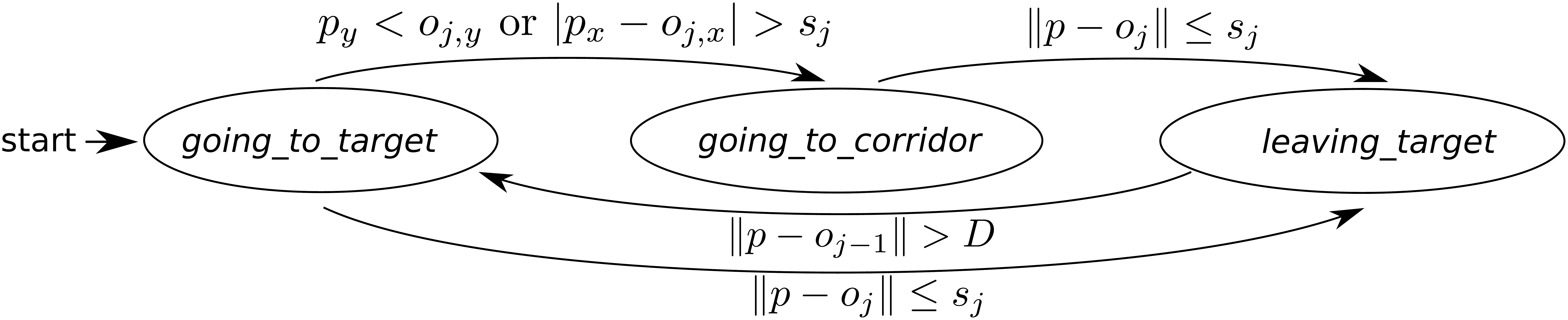}
  \caption{State machine transitions for SQF algorithm.}
  \label{fig:statesSQF}
\end{figure}

Algorithm \ref{alg:sqf} presents the SQF algorithm. Robots have the states \emph{go\-ing\-\_\-to\-\_\-tar\-get}, \emph{leav\-ing\-\_\-tar\-get} and \emph{go\-ing\-\_\-to\-\_\-cor\-ri\-dor}, which respectively means the robot is going straightly to the target region, it is leaving it, and it is going to the corridor. The robots begin at \emph{go\-ing\_to\_tar\-get} state. Figure \ref{fig:statesSQF} shows the state machine and its transitions, assuming that the robot may have two or more targets located at $o_{j} = (o_{j,x}, o_{j,y})$, the circular target region has radius $s_{j}$, and $j$ is its current target index. Note that this algorithm is only executed when the robot is at a distance at most $D$ from the target centre.

\begin{algorithm}
  \setstretch{1.2}
  \SetKwInOut{Input}{Input}
  \SetKwInOut{Output}{Output}
  \Input{$K_{SQF}:$ force magnitude; \newline
    $D:$ the length of the corridor; \newline
    $o_{1} = (o_{1,x},o_{1,y}), \dots, o_{n} = (o_{n,x},o_{n,y}):$ a list of $n \ge 2$ circular target region centres; \newline
    $s_{1}, \dots, s_{n}$ a list of $n$ circular target region radii;  \newline
    $j:$ the current target index; \newline
    $I_{min}, I_{d}:$ minimum and default influence radius of the robot. 
  }
  \Output{$F=(F_{x},F_{y}):$ force vector; \newline
    $I:$ influence of the robot for repulsive force calculation. 
  }
  Get position of the robot $p=(p_{x},p_{y})$, and let $o = o_{j} = (o_{x},o_{y})$ and $s = s_{j}$\;
  \uIf{$\|p-o\| \le s$}{
    state $\leftarrow leav\-ing\-\_\-tar\-get$\; 
    Increment $j$, then let $o = o_{j} = (o_{x},o_{y})$ and $s = s_{j}$\; 
  }
  \uIf{state $\neq$ leav\-ing\-\_\-tar\-get}{
    \uIf {$\|p - o\| \le D$ and ($p_{y} < o_{y}$ or $|p_{x} - o_{x}| > s$)}{
      state $\leftarrow go\-ing\_to\_cor\-ri\-dor$\;
      Set $F$ according to (\ref{eq:vortex11}) and (\ref{eq:vortex12})\;
    }
    \uElse{
      state $\leftarrow go\-ing\_to\_tar\-get$\;
      F $\leftarrow K_{SQF}\frac{o-p}{\|o - p\|}$\;
    }  
  }
  \uElse{
    \uIf{$\|p - o_{j-1}\| \le D$}{
       Set $F$ according to (\ref{eq:vortex21}) and (\ref{eq:vortex22})\;
    }
    \uElse{
      state $\leftarrow go\-ing\_to\_tar\-get$\;
      F $\leftarrow K_{SQF}\frac{o-p}{\|o - p\|}$\;
    }
  }
  \uIf{state = go\-ing\_to\_tar\-get or state = leav\-ing\-\_\-tar\-get}{
    $I \leftarrow I_{min}$\;
  }
  \uElseIf{state = go\-ing\_to\_cor\-ri\-dor and $p_{y} > o_{y}$ and $|p_{x} - o_{x}| < I_{d} - I_{min}$}{
    $I \leftarrow I_{min} + |p_{x} - o_{x}|$\;
  }
  \uElse{
    $I \leftarrow I_{d}$\;
  }
  \Return $(F,I)$\;
  \caption{SQF algorithm, executed at every update in the position of the robot.}
  \label{alg:sqf}
\end{algorithm}

\subsection{Touch and Run Vector Fields}

\begin{figure}[t]
  \centering
  \subfloat[]{
    \includegraphics[width=0.47\columnwidth]{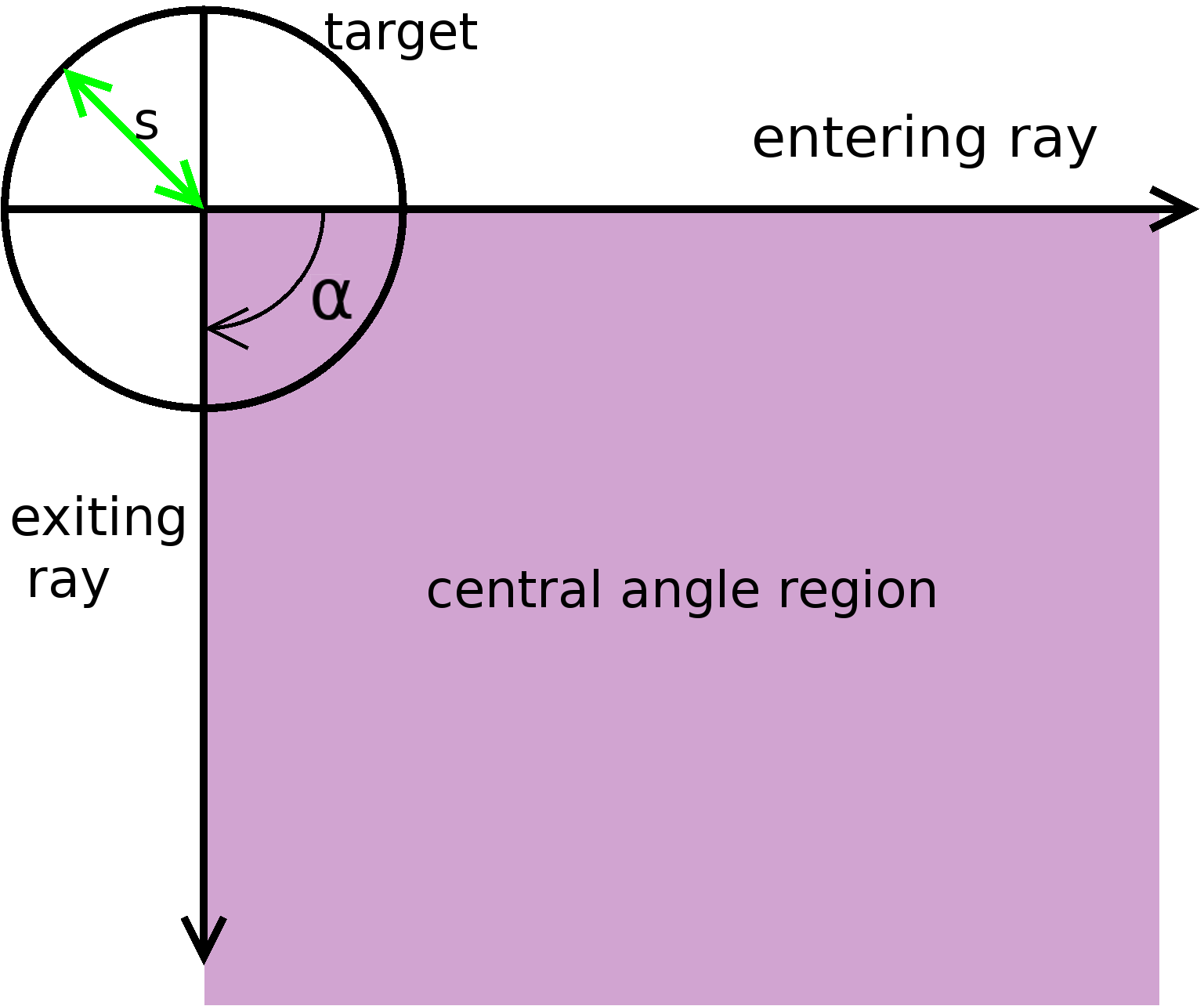}
  }
  \subfloat[]{
    \includegraphics[width=0.47\columnwidth]{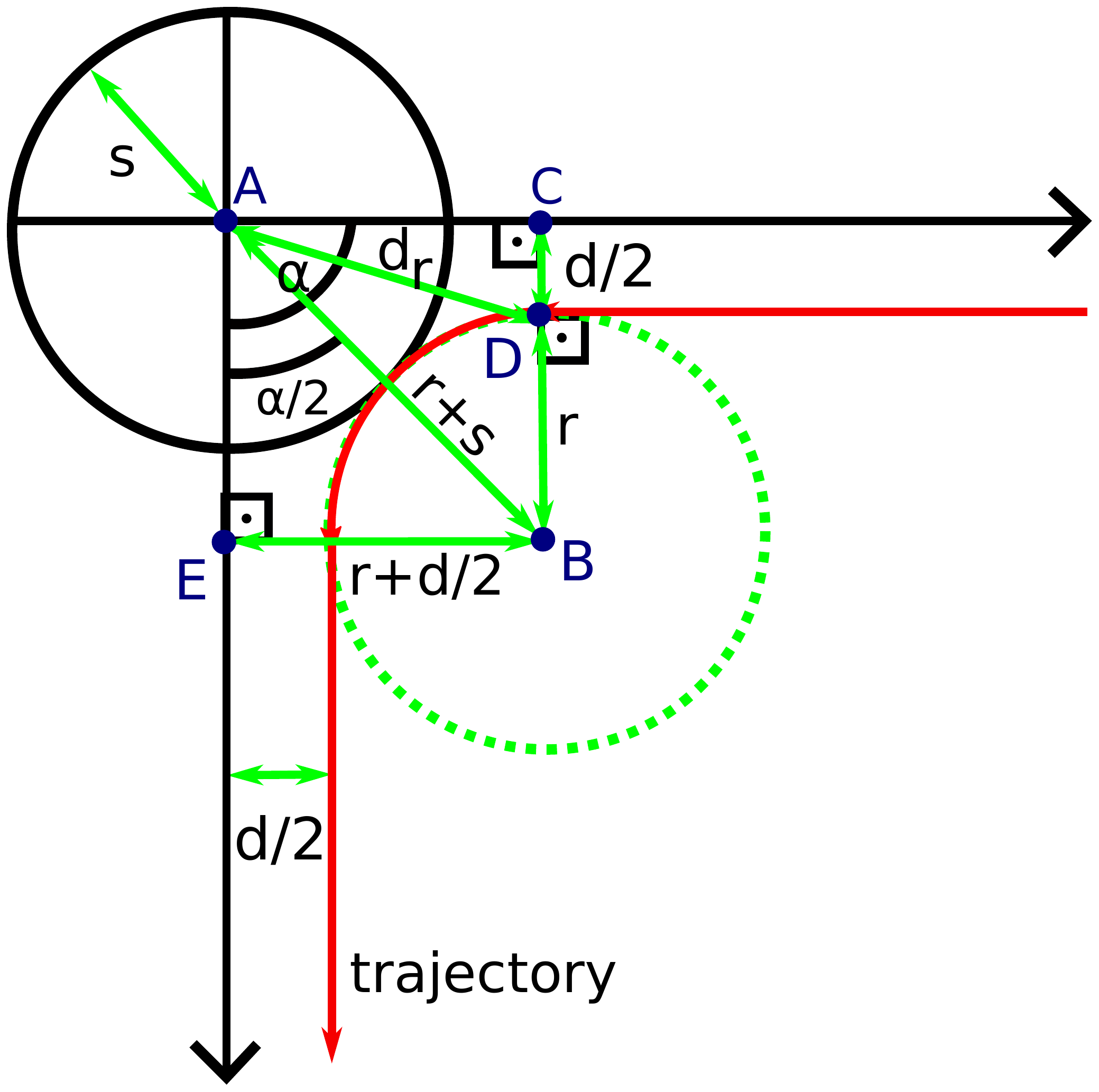}
  }
  \caption{Illustration of the touch and run strategy (from \citep{arxivTheory}). (a) Central angle region and its exiting and entering rays, defined by the angle $\alpha$. (b) The trajectory of a robot next to the target, in red. Here we have the relationship between the target area radius ($s$), the minimum safety distance between the robots ($d$), the turning radius ($r$), the central region angle ($\alpha$) and the distance from the target centre for a robot to begin turning ($d_{r}$). The green dashed circle represents the whole turning circle.}
  \label{fig:theoretical:central_angle_linka}
\end{figure}

Since a robot should spend as little time as possible near the target, in \citep{arxivTheory}, we imagined a simple scenario where robots travel in predefined curved lanes and tangent to the target area, where they spend minimum time on the target. 
To avoid collisions with other robots, the trajectory of a robot nearby the target is circular,
and the distance between each robot must be at least $d$ at any part of the trajectory. 
Hence, no lane crosses another, and each lane occupies a region defined by an angle in the target area -- the central region angle -- that we denote by $\alpha$, shown in Figure \ref{fig:theoretical:central_angle_linka} (a). Figure \ref{fig:theoretical:central_angle_linka} (b) shows the trajectory of a robot towards the target region following that \emph{touch and run} strategy. The robot first follows the boundary of the central angle region -- that is, the entering ray -- at a distance of $d/2$.
Then, it arrives at a distance of $s$ of the target centre using a circular trajectory with a turning radius $r$. 
As the trajectory is tangent to the target shape, it is close enough to consider that the robot reached the target region.
Finally, the robot leaves the target by following the second boundary of the central angle region -- that is, the exiting ray -- at a distance of $d/2$. 

Based on this strategy, we devise an algorithm to mimic the curved trajectories around the target region using vector fields, named Touch and Run Vector Fields (TRVF). The vector fields are only used inside a circle with radius $D$ around the target centre and apply forces on the robot to make it follow a circular trajectory similar to the one described in the touch and run strategy.

As a way to do that, we adapted and combined the vector fields proposed by \citet{vectorField2006} for straight line and orbit following trajectories. We use their straight line following vector fields whenever the robot has to follow a straight trajectory and an anti-clockwise orbit following vector field for curved trajectories. We will first explain our modified functions for these trajectories, and afterwards, we will introduce the TRVF algorithm.

Our adaptation of their work concerns reducing the orbit following to a circular sector, instead of the full circle, and adding an attractive force towards a fixed point when the robots are performing the orbit following next to the target region. The reduction to a circular sector is due to the curved trajectory next to the target having $\alpha < 2\pi$. We add attractive forces towards a fixed point to the orbit following vector field intending to avoid the repulsive forces caused by nearby robots to push a robot far away from that fixed point.

\begin{figure}[h]
  \centering 
  \subfloat[]{\includegraphics[width=0.375\linewidth]{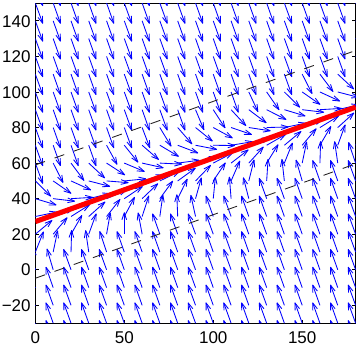}}\,
  \subfloat[]{\includegraphics[width=0.39\linewidth]{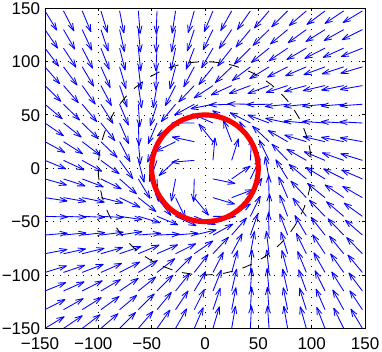}}
  \caption{Straight line and orbit following vector fields (from \citep{vectorField2006}).}
  \label{fig:vforigpaper}
\end{figure}

Figure \ref{fig:vforigpaper} shows the straight line and orbit following vector fields proposed by \citet{vectorField2006}. Algorithm \ref{alg:straightlinevf} presents our adapted function that computes the straight line following vector field, its parameters and values returned. It assumes a fixed segment to follow with initial and final waypoints $w_{i}$ and $w_{f}$, respectively. Based on the current position and orientation of the robot, $p$ and $\xi$, respectively, a force vector with magnitude $K_{TRVF}$ is computed. The orientation of the force vector changes gradually as the robot approaches the segment. These changes are calculated using the maximum linear speed of the robot $v$, and its rate of change depends on the constants $k_{s}$ and $K_{r}$.

\begin{algorithm}[t!]
  \setstretch{1.2}
  \SetKwFunction{Fn}{Function}
  \SetKwInOut{Input}{Input}
  \SetKwInOut{Output}{Output}
  \Fn straightPathFollowing ($K_{TRVF}, p,\xi, w_{i}, w_{f}, I_{d}, k_{s}, K_{r}, v$){
    
    \Input{$K_{TRVF}:$ force magnitude; \newline
      $p = (p_{x},p_{y}), \xi:$ current position and orientation of the robot; \newline
      $w_{i} = (w_{i,x},w_{i,y}), w_{f} = (w_{f,x},w_{f,y}):$ initial and final waypoints for the segment to follow; \newline
      $I_{d}:$ influence radius of the robot; \newline
      $k_{s} > 1$: constant for exponentiation in the vector field calculation; \newline
      $K_{r}:$ constant for proportional angular speed controller; \newline
      $v$: maximum linear speed.
    }
    \Output{$F=(F_{x},F_{y}):$ force vector; \newline
      $t:$ indicates the position along the path (0 when the position of the robot is on the line perpendicular to $\overline{w_{i}w_{f}}$ on the initial waypoint and 1 when that happens for the final waypoint). 
    }
    $w_{fi} = w_{f} - w_{i}$\;
    $t \leftarrow \frac{(p-w_{i}) \cdot w_{fi}}{\|w_{fi}\|^{2}}$\; 
    \lIf {$t \ge 1$} {\Return $((0,0),t)$\label{line:noforcestraight}}
    $\xi_{f} \leftarrow \atantwo(w_{fi,y},w_{fi,x})$\;
    $\epsilon \leftarrow \|p - w_{i} - t w_{fi}\|$\;
    $\rho \leftarrow sign(w_{fi} \times (p-w_{i}))$ \label{line:signcross}\;
    $\tau \leftarrow I_{d}/5$\;
    $\xi_{e} \leftarrow \pi/2$\;
    \uIf{$\epsilon > \tau$}{
      $\xi_{c} \leftarrow \xi_{f} - \rho \xi_{e}$\;
    }
    \uElse{
        $\epsilon \leftarrow \rho\epsilon$\;
        $ P_{1} \leftarrow \left(\frac{\epsilon}{\tau}\right)^{k_{s}}$; 
        \lIf {$P_{1}$ is NaN} {$P_{1} \leftarrow 0$\label{line:nancheck1}}
        $ P_{2} \leftarrow \epsilon^{k_{s}-1}$; 
        \lIf {$P_{2}$ is NaN} {$P_{2} \leftarrow 0$\label{line:nancheck2}}
        $\xi_{c} \leftarrow \xi_{f} - \xi_{e} P_{1} - \left(\frac{k_{s} \xi_{e} v}{K_{r}\tau^{k_{s}}}\right)P_{2} \sin(\xi)$\;
    }
    $F \leftarrow K_{TRVF}(\cos(\xi_{c}),\sin(\xi_{c}))$\;
    \Return $(F,t)$\;
  }
  \caption{Straight-line vector field algorithm adapted from \citet{vectorField2006}.}
  \label{alg:straightlinevf}
\end{algorithm}

In the original algorithm, $\xi_{e}$ is the entry heading angle, and $\tau$ is the distance perpendicular to the line from which the vector field makes the heading angle begin to change (in Figure \ref{fig:vforigpaper} (a), it is the distance from the dashed line to the solid line). In our adaptation, we set $\tau = I_{d}/5$ and $\xi_{e} = \pi/2$, for a given default influence radius of the robot  $I_{d}$. Additionally, we only compute the force vector if the robot did not reach the point $w_{f}$ or if the robot is in a position on the plane perpendicular to $\overline{w_{i}w_{f}}$ limited by its endpoints,  otherwise, we return force vector zero (line \ref{line:noforcestraight}). At line \ref{line:signcross}, $sign$ returns $-1$ for negative numbers and $1$, otherwise. Here $\times$ stands for the cross-product of two vectors. We also add a verification for not a number (NaN) in the exponentiations and replace the result for zero as it is a common source of numerical errors (lines \ref{line:nancheck1} and \ref{line:nancheck2}).  We are assuming a proportional angular speed controller such that if the robot has orientation $\xi$ and needs to turn to orientation given by $\xi_{c}$, then it uses angular speed $K_{r}(\xi_{c} - \xi)$ for a chosen $K_{r} > 0$.

Algorithm \ref{alg:circularvf} shows our modified function that computes the orbit following vector field, its parameters and values returned. Let the orbit to follow be centred at a point $c = (c_{x},c_{y})$ and have radius $R$. As in the straight line following algorithm, given the current position and orientation of the robot, $p$ and $\xi$, respectively, a force vector with magnitude $K_{TRVF}$ is calculated. In this work, we intend to use the orbit following algorithm to perform curves, so a waypoint $w_{f}$ must be given to indicate where the orbit following must stop. As in the previous algorithm, the orientation of the force vector changes gradually as the robot approaches the orbit, and these changes are computed using the maximum linear speed of the robot $v$, and its rate of change depends on the constants $k_{o}$ and $K_{r}$.

\begin{algorithm}[t!]
  \setstretch{1.2}
  \SetKwFunction{Fn}{Function}
  \SetKwInOut{Input}{Input}
  \SetKwInOut{Output}{Output}
  \Fn orbitPathFollowing ($K_{TRVF}, c, R, p, \xi, w_{f}, k_{o}, K_{r}, v$){

    \Input{$K_{TRVF}:$ force magnitude; \newline
      $c, R:$ centre and radius for orbit following; \newline
      $p = (p_{x},p_{y}), \xi:$ current position and orientation of the robot; \newline
      $w_{f} = (w_{f,x},w_{f,y}):$ final waypoint; \newline
      $k_{o} > 1$: constant for exponentiation in the vector field calculation; \newline
      $K_{r}:$ constant for proportional angular speed controller; \newline
      $v$: maximum linear speed.
    }
    \Output{$F=(F_{x},F_{y}):$ force vector; \newline
      $t:$ indicates the position along the path (positive when the robot did not cross $\overrightarrow{cw_{f}}$). 
    }
    $q \leftarrow p - c$\; 
    $t \leftarrow q \times (w_{f} - c)$\;
    \lIf{$t \le 0$} {\Return $((0,0),t)$\label{line:noforceorbit}}
    $\gamma \leftarrow \atantwo(q_{x},q_{y})$\label{line:gammacircle}\; 
    \uIf{$\|q\| > 2R$}{
      $\xi_{c} \leftarrow \gamma - \frac{5\pi}{6} + \frac{v}{\|q\|}\sin(\xi - \gamma)$\;
    }
    \uElse{
       $ P_{1} \leftarrow \left(\frac{\|q\|-R}{R}\right)^{k_{o}}$; 
       \lIf {$P_{1}$ is NaN} {$P_{1} \leftarrow 0$}
       $ P_{2} \leftarrow (\|q\|-R)^{k_{o}-1}$; 
       \lIf {$P_{2}$ is NaN} {$P_{2} \leftarrow 0$}
       $\xi_{c} \leftarrow \gamma  - \frac{\pi}{2} - \frac{\pi}{3} P_{1} - \frac{v}{K_{r}\|q\|}\sin(\xi - \gamma) - \frac{k_{o} v \pi}{3 R^{k_{o}} K_{r}} P_{2} \cos(\xi-\gamma)$\;
    }
    $F \leftarrow K_{TRVF}(\cos(\pi/2 - \xi_{c}),\sin(\pi/2 - \xi_{c}))$\label{line:Fturnback}\; 
    \Return $(F,t)$\;
  }
  \caption{Anti-clockwise orbit following vector field algorithm adapted from \citet{vectorField2006}.}
  \label{alg:circularvf}
\end{algorithm}

This algorithm only generates a force vector different from zero if the robot did not cross  $\overrightarrow{cw_{f}}$ for a given waypoint $w_{f}$. We verify if it crossed by the positive sign of the cross product between the vector $q = p - c$ (that is, the vector from the centre of the orbit to the position $p$ of the robot) and the vector $w_{f} - c$. As the orbit is anti-clockwise, the orbit following vector field makes the cross product $q \times (w_{f} - c)$ decrease with time, being zero when $q$ has the same orientation as $w_{f} - c$ and negative when the orientation of $q$ is greater than the orientation of $w_{f} - c$. We return force vector zero when the robot is in a position beyond $\overrightarrow{cw_{f}}$ or intersects this ray (line \ref{line:noforceorbit}).

\begin{figure}[t]
  \begin{minipage}[b]{0.39\linewidth}
    \centering 
    \includegraphics[width=0.65\linewidth]{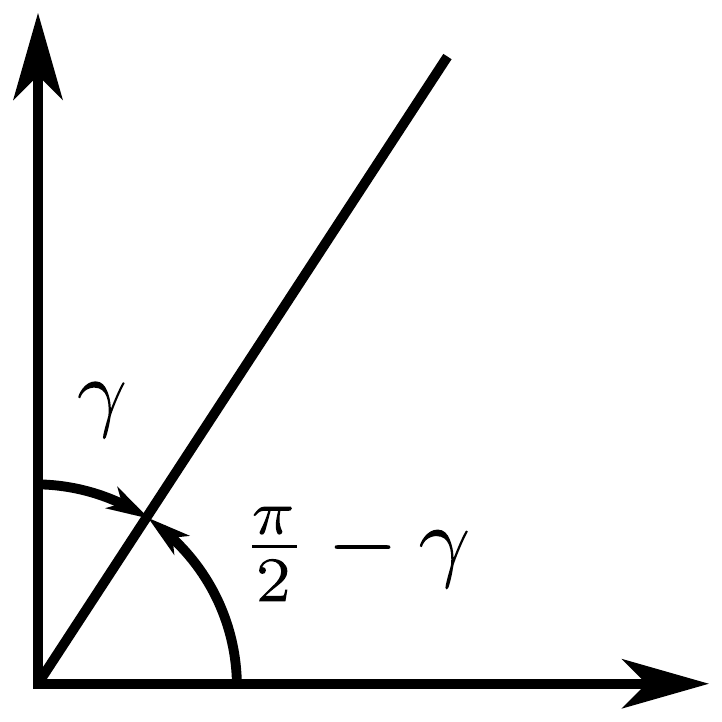}
    \caption{If a line has angle $\gamma$ with the $y$-axis assuming clockwise orientation, then this same line has angle $\pi/2-\gamma$ with the $x$-axis using anti-clockwise orientation, as the $y$-axis makes an angle of $\pi/2$ with the $x$-axis.}
    \label{fig:xcorient}
  \end{minipage}\quad
  \begin{minipage}[b]{0.6\linewidth}  
    \centering 
    \includegraphics[width=0.76\linewidth]{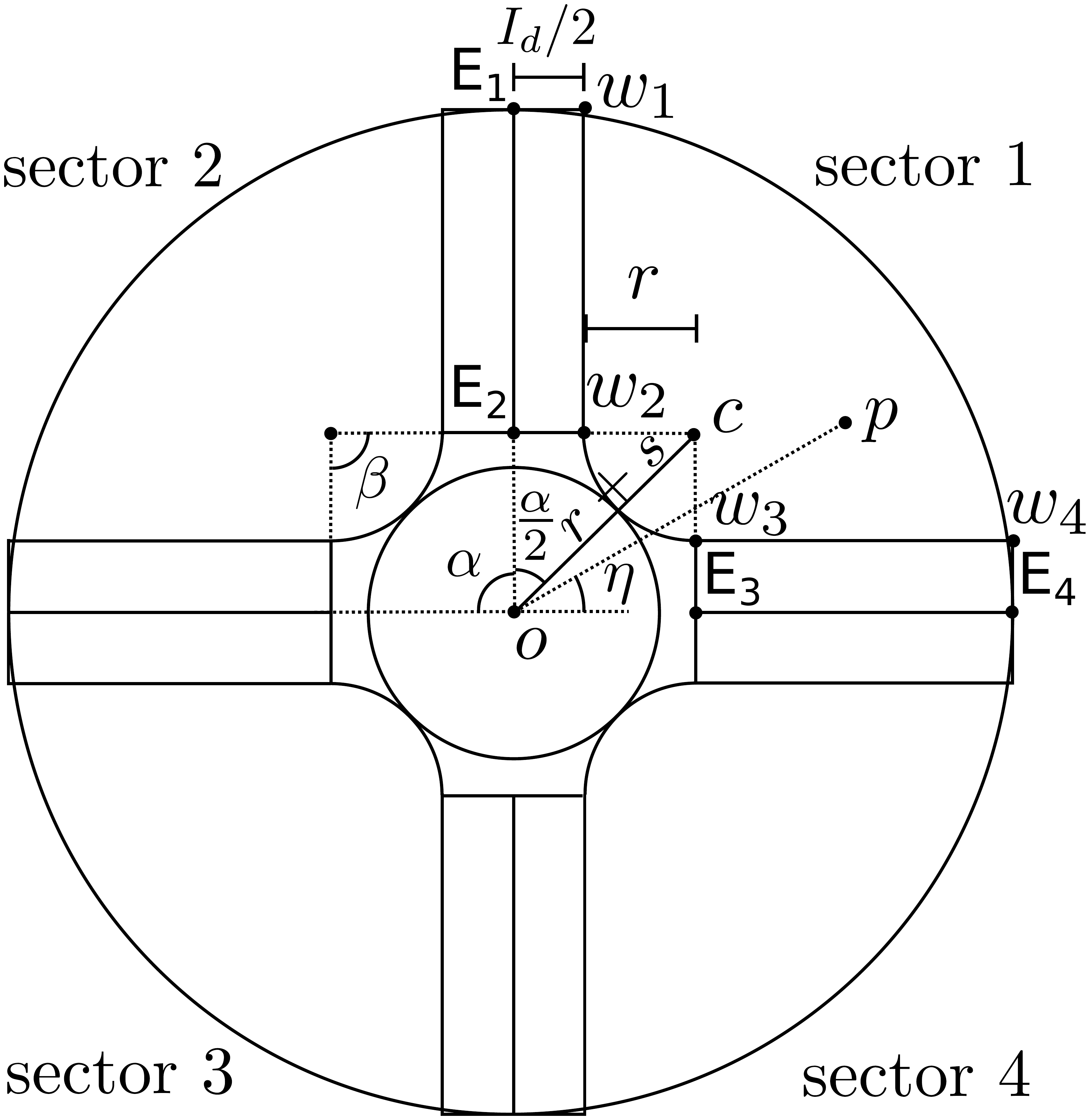}
    \caption{The robots follow the waypoints $w_{1}$, $w_{2}$, $w_{3}$ and $w_{4}$ depending on the sector where they are located. The centre $c$ for the curved trajectory between waypoints $w_{2}$ and $w_{3}$ is distant by $r+s$ from the target centre $o$.  In this example, we have $K=4$ and a robot at position $p$ and at sector 1.}
    \label{fig:waypointsTRVF}
  \end{minipage}
\end{figure}

At line \ref{line:gammacircle}, $\gamma$ is the heading from the centre of the orbit to the position of the robot. \citet{vectorField2006} use clockwise orientation and they assume null orientation when the robot is at the $y$-axis, thus we are using $\gamma = \atantwo(q_{x},q_{y})$ instead of $\atantwo(q_{y},q_{x})$ as in the original algorithm. Due to this, at line \ref{line:Fturnback} we use $\pi/2 - \xi_{c}$ for computing the result using the usual orientation, that is, anti-clockwise with null orientation at the $x$-axis, as justified by Figure \ref{fig:xcorient}.

As presented in the touch and run strategy of \cite{arxivTheory}, the TRVF algorithm uses $K$ lanes. Each robot calculates four waypoints to follow in the lane located on the right side of the robot, assuming it is facing the target located at point $o$. Figure \ref{fig:waypointsTRVF} shows these four waypoints, which are based on the position of the robot $p$, and must be followed in order. The four waypoints are calculated depending on the sector where they are. There are $K$ sectors, one per lane, numbered from 1 to $K$. Let $\eta$ be the angle of the vector $p-o$. Thus, a sector $i$ of a robot is calculated by 
$
i = \left\lfloor\frac{\eta}{\alpha}\right\rfloor + 1.
$

Let $E_{1}$ be the point in the entering ray on the boundary of the circle centred at $o$ with radius $D$, $E_{2}$ be the point in the entering ray perpendicular to the point where the robot starts to turn in the touch and run strategy (as presented in Figure \ref{fig:theoretical:central_angle_linka} (b)), $E_{3}$ be the point in the exiting ray perpendicular to the point where the robot ends the turn in the touch and run strategy, and $E_{4}$ be the point in the exiting ray on the boundary of the circle centred at $o$ with radius $D$. In Figure \ref{fig:waypointsTRVF}, these points are exemplified for sector 1. For a given sector $i$, the entering and exiting rays will have angle $i\alpha$ and $(i-1)\alpha$ with the $x$-axis, respectively. The position vector $w_{1}$ is given by the sum of the vector $o$, the vector going from $o$ to $E_{1}$ and the vector of modulus $I_{d}/2$ from $E_{1}$ rotated $\pi/2$ anti-clockwise from the entering ray. Thus,

\begin{equation}
  \begin{aligned}
    w_{1} 
      &= o + (E_{1} - o) + \frac{I_{d}}{2}(\cos(i\alpha-\pi/2),\sin(i\alpha-\pi/2)) \\
      &= o + D(\cos(i\alpha),\sin(i\alpha)) + \frac{I_{d}}{2}(\sin(i\alpha),-\cos(i\alpha)). 
  \end{aligned}
  \label{eq:w1}
\end{equation}

The vector $w_{2}$ is given similarly to $w_{1}$ but using $E_{2}$. The distance from $o$ to $E_{2}$ is $\sqrt{(r+s)^{2} - (r+I_{d}/2)^{2}}$, because this is the same as $\overline{AC}$ in the Figure \ref{fig:theoretical:central_angle_linka} (b), but using $d$ instead of $I_{d}$. In other words, in this figure, we have the distance $d_{r}$ from the target centre where the robots begin turning.  By symmetry, this is the same distance from the target centre where the robots stop turning. From the right triangle $ABC$ on that figure, we have $|\overline{AC}| = \sqrt{(r+s)^{2} - (r+d/2)^{2}}$.

The value of $r$ can be calculated from Figure \ref{fig:theoretical:central_angle_linka} (b) replacing $d$ by $I_{d}$ as well. That is, we can see that the right triangle ABE has angle $\widehat{EAB} = \alpha/2$,  hypotenuse $r + s$ and cathetus $r + d/2$. Hence, it directly follows that 

\begin{equation}
  sin(\alpha / 2) = \frac{r + d/2}{r + s}
  \Rightarrow
  r = \frac{s   \sin(\alpha / 2) - I_{d}/2}{1 - \sin(\alpha / 2)},
  \label{eq:relationangles}
\end{equation}
after reordering and replacing $d$ by $I_{d}$. Then, 

\begin{equation}
  \begin{aligned}
    w_{2} 
      &= o + (E_{2} - o) + \frac{I_{d}}{2}(\cos(i\alpha-\pi/2),\sin(i\alpha-\pi/2)) \\
      &= o + \sqrt{(r+s)^{2} - (r+I_{d}/2)^{2}}(\cos(i\alpha), \\
      &\phantom{=} \sin(i\alpha)) + \frac{I_{d}}{2}(\sin(i\alpha),-\cos(i\alpha)). 
  \end{aligned}
\end{equation}

We use similar reasoning to calculate $w_{3}$ and $w_{4}$, however, for the exiting ray, we use the angle $(i-1)\alpha$ instead of $i\alpha$. Thus,

\begin{equation}
  \begin{aligned}
    w_{3} 
      &= o + (E_{3} - o) + \frac{I_{d}}{2}(\cos((i-1)\alpha-\pi/2),\sin((i-1)\alpha-\pi/2)) \\
      &= o + \sqrt{(r+s)^{2} - (r+I_{d}/2)^{2}}(\cos((i-1)\alpha),\sin((i-1)\alpha)) +\\
      &\phantom{=\ } \frac{I_{d}}{2}(\sin((i-1)\alpha),-\cos((i-1)\alpha)), 
  \end{aligned}
\end{equation}
and

\begin{equation}
  \begin{aligned}
    w_{4} 
      &= o + (E_{4} - o) + \frac{I_{d}}{2}(\cos((i-1)\alpha-\pi/2),\sin((i-1)\alpha-\pi/2)) \\
      &= o + D(\cos((i-1)\alpha),\sin((i-1)\alpha)) +\frac{I_{d}}{2}(\sin((i-1)\alpha),\\
      &\phantom{=\ } -\cos((i-1)\alpha)). 
  \end{aligned}
\end{equation}

For the sector $i$, the curve with radius $r$ has centre $c$, which is distant by $r+s$ from the target centre. The vector $c - o$ has orientation $i\alpha - \frac{\alpha}{2}$ (Figure \ref{fig:waypointsTRVF}), so 

\begin{equation}
  c = o + (r + s)\left(\cos\left(\left(i - \frac{1}{2}\right)\alpha\right), \sin\left(\left(i - \frac{1}{2}\right)\alpha\right)\right).
  \label{eq:cisectori}
\end{equation}

In order to keep track of the path following, we also use a state machine. Every robot has six states. 
The initial state is \emph{go\-ing\-\_\-to\-\_\-tar\-get}. 
The robot remains in this state until it reaches the distance $D$ from the target, then it changes to \emph{go\-ing\-\_\-to\-\_\-en\-trance\-\_\-straight\-\_\-path}. In this state, the robot follows an orbit following vector field centred at the target centre $o$ and radius $D$ using Algorithm \ref{alg:circularvf} with the parameter  $w_{f} = w_{1}$. The robot changes to the state \emph{on\-\_\-en\-trance\-\_\-straight\-\_\-path} when the Algorithm \ref{alg:circularvf} outputs $t \le 0$. This state indicates that the robot is following the straight-line vector field towards the target region until it reaches a position after the line orthogonal to the following line at the waypoint $w_{2}$. This happens when the variable $t$ returned by the Algorithm \ref{alg:straightlinevf} is greater than or equal to 1. 

When this occurs, the robot changes to \emph{on\-\_\-en\-trance\-\_\-curved\-\_\-path} and is impelled by the sum of two forces: the orbit following force for the curved trajectory with centre at $c$ and radius $r$ and an attraction force towards the target centre $o$. The second force was added because, next to the target region, as the number of robots grows, repulsive forces may push a robot on state \emph{on\-\_\-en\-trance\-\_\-curved\-\_\-path} far from the target. The orbit following force field by itself does not attract to the target region, so the other force is a counter-effect to this pushing. However, the attractive force magnitude must be higher than the orbit following force magnitude to avoid their vector sum being null. Consequently, we use fixed norm $1.5 K_{TRVF}$ for that attractive force, sum these two vectors, then normalise the result to $K_{TRVF}$. For the Algorithm \ref{alg:circularvf} in this state we use $w_{f} = w_{3}$.

When the robot reaches the target region, it changes to \emph{on\-\_\-ex\-it\-\_\-curved\-\_\-path}. In this state, the robot continues to follow the previous orbit following but we add an attractive force towards the waypoint $w_{3}$ by a similar normalised summation as described for the previous state. The transition to the next state,  \emph{on\-\_\-ex\-it\-\_\-straight\-\_\-path}, is done when Algorithm \ref{alg:circularvf} returns $t \le 0$. Then, the robot follows the straight line following force field until it leaves the circle with radius $D$ around the target centre, changing back to \emph{go\-ing\-\_\-to\-\_tar\-get} and going to the next target region. We also added a transition from the state \emph{on\-\_\-ex\-it\-\_\-curved\-\_\-path} to \emph{go\-ing\-\_\-to\-\_tar\-get} because $D$ could be not much greater than $s$, so the robot could leave the circle with radius $D$ due to repulsive forces before making a transition to \emph{on\-\_\-ex\-it\-\_\-straight\-\_\-path}. 

\begin{figure}[h]
  \centering 
  \subfloat[State machine]{\includegraphics[width=0.46\linewidth]{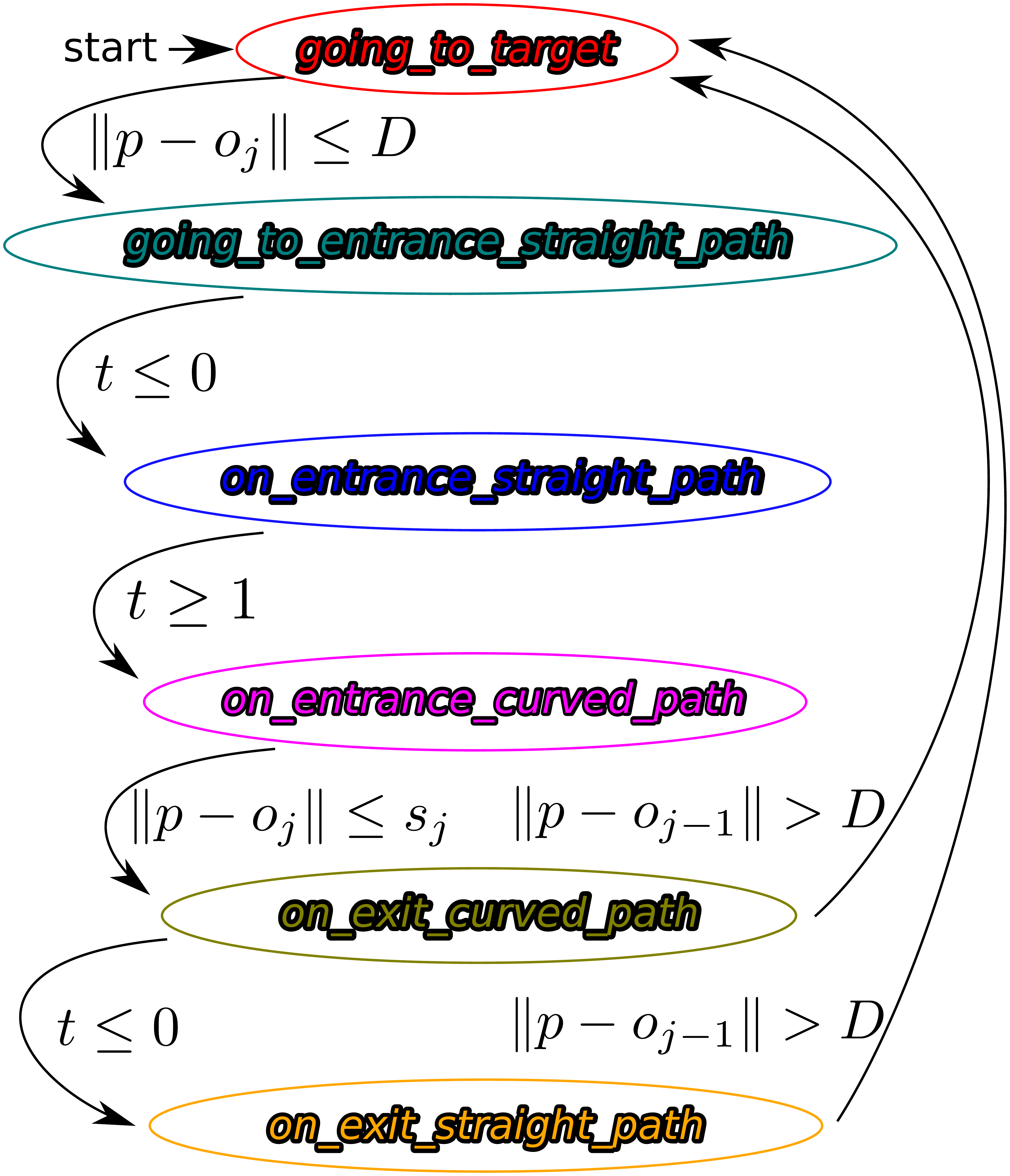}}
  \subfloat[Vector field]{\includegraphics[width=0.54\linewidth]{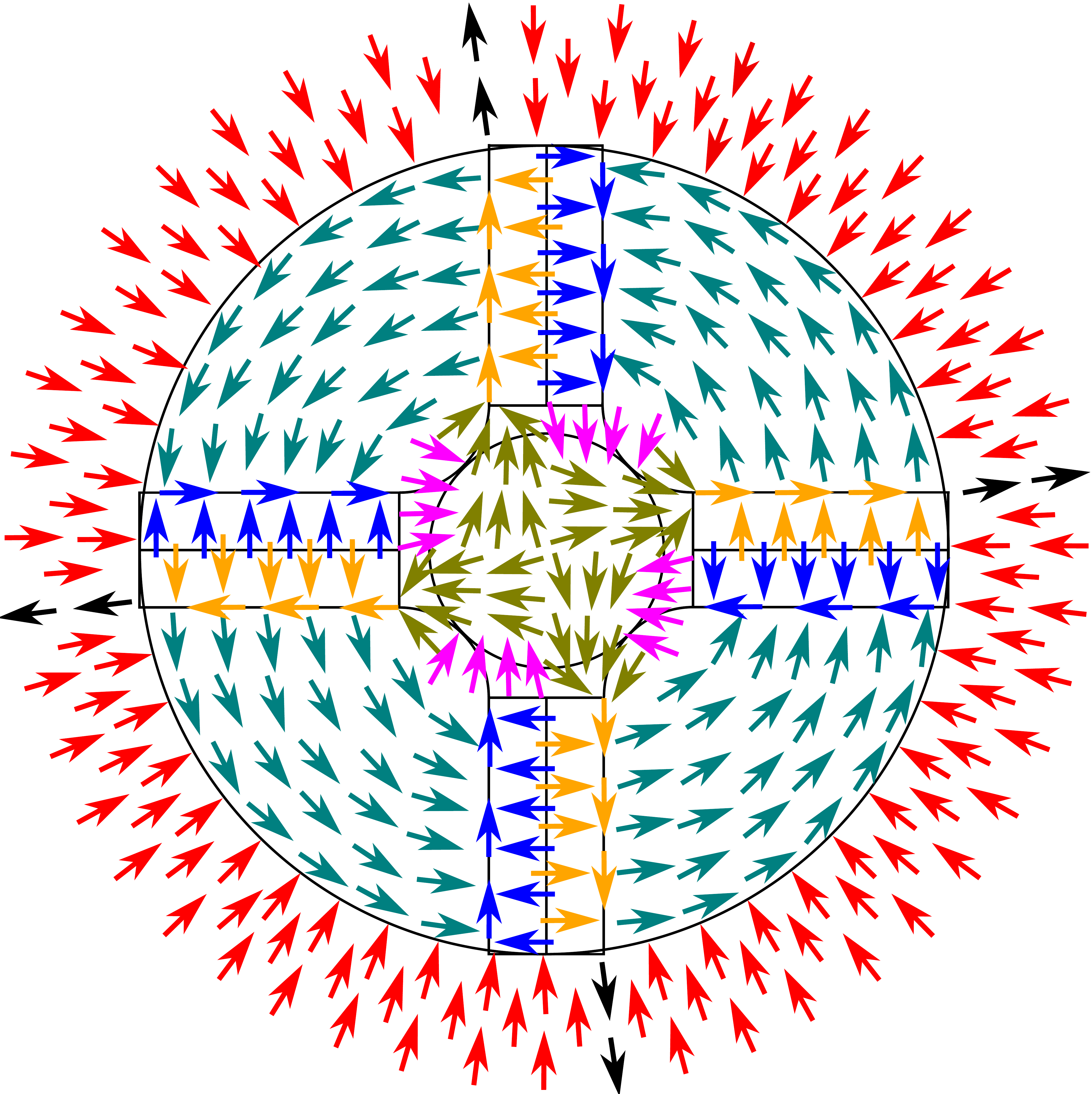}}
  \caption{Vector field and state machine for the TRVF algorithm. The colours in the vector field represent the expected robot state at the shown vector positions. The black vector indicates the effect of the repulsive force from the target region when the robot is more distant than $D$ from the target already reached.  The inner and outer circles have a radius of $s$ and $D$, respectively. Here, we are using $K=4$.} 
  \label{fig:statemachinevectorsTRVF}
\end{figure}

As the robot is attracted to the next target when it is on \emph{go\-ing\-\_\-to\-\_tar\-get}, depending on its position, it could be impelled to cross the previous target region again. Thus, we apply a repulsive force from the circle with radius $D$ and centre at the previous target centre as if it was a huge obstacle. This makes the robots avoid the previous target region giving space to other robots entering or exiting. Figure \ref{fig:statemachinevectorsTRVF} (a) summarises these transitions and Figure \ref{fig:statemachinevectorsTRVF} (b) shows the vector field and the expected state for a robot at the displayed positions assuming no influence of the repulsive force of the other robots.

Finally, Algorithm \ref{alg:trvf} presents the TRVF algorithm.  The condition at line \ref{line:waypointscalculated} can be checked by a global boolean variable initialised as false before the robot goes to the next target. Without this condition, the robots would change lanes if they get pushed to another one, resulting in more congestions. At lines \ref{line:repfun1} and \ref{line:repfun2}, we assume any chosen repulsive force function.

\begin{algorithm}[t!]
  \setstretch{1.2}
  \SetKwInOut{Input}{Input}
  \SetKwInOut{Output}{Output}
  \Input{
    $K_{TRVF}:$ force magnitude; \newline
    $K$: number of lanes; \newline
    $D:$ the length of the corridor; \newline
    $o_{1} = (o_{1,x},o_{1,y}), \dots, o_{n} = (o_{n,x},o_{n,y}):$ a list of $n \ge 2$ circular target region centres; \newline
    $s_{1}, \dots, s_{n}$ a list of $n$ circular target region radii;  \newline
    $j:$ the current target index; \newline
    $I_{d}:$ influence of the robot; \newline 
    $k_{s} > 1, k_{o} > 1$: constant for exponentiation in the straight line following and orbit following vector fields, respectively; \newline 
    $K_{r}:$ constant for proportional angular speed controller; \newline
    $v$: maximum linear speed.
  }
  \Output{ 
    $F=(F_{x},F_{y}):$ force vector;
  }
  Get position of the robot $p=(p_{x},p_{y})$, its orientation $\xi$, and let $o = o_{j} = (o_{x},o_{y})$ and $s = s_{j}$\;
  \uIf{$w_{1}, \dots, w_{4},$ and $c$ were not calculated for target index $j$}{ \label{line:waypointscalculated}
    $ i \leftarrow \left\lfloor\frac{\atantwo(p_{y} - o_{y},p_{x} - o_{x})}{2\pi/K}\right\rfloor + 1$\;
    Use Eqs. \ref{eq:w1}-\ref{eq:cisectori} to calculate $w_{1}, \dots, w_{4}, r$ and $c$\; 
  }
  \uIf {state = go\-ing\-\_\-to\-\_tar\-get}{
    \lIf {$\|p-o\| \le D$}{state $\leftarrow go\-ing\-\_\-to\-\_\-en\-trance\-\_\-straight\-\_\-path$}
    \uElse{
      $F \leftarrow K_{TRVF}\frac{o-p}{\|o - p\|}$\;
      $F_{R} \leftarrow$ repulsiveForceFromTarget($o_{j-1},D$)\label{line:repfun1}\; 
      \Return $K_{TRVF}\frac{F + F_{R}}{\|F + F_{R}\|}$\; 
    }
  }
  \uIf {state = go\-ing\-\_\-to\-\_\-en\-trance\-\_\-straight\-\_\-path}{
    $(F,t) \leftarrow orbitPathFollowing (K_{TRVF}, o, D, p, \xi, w_{1}, k_{o}, K_{r}, v)$\;
    \lIf {$t \le 0$}{state $\leftarrow on\-\_\-en\-trance\-\_\-straight\-\_\-path$}
    \lElse {\Return $F$}
  }
  \caption{TRVF algorithm (continues on Algorithm \ref{alg:trvfcont}).}
  \label{alg:trvf}
\end{algorithm}

\begin{algorithm}[t!]
  \setstretch{1.2}
  \setcounter{AlgoLine}{15}
  \uIf{state = on\-\_\-en\-trance\-\_\-straight\-\_\-path}{
    $(F,t) \leftarrow straightPathFollowing (K_{TRVF}, p, \xi, w_{1}, w_{2}, I_{d}, k_{s}, K_{r}, v)$\;
    \lIf {$t \ge 1$}{state $\leftarrow on\-\_\-en\-trance\-\_\-curved\-\_\-path$}
    \lElse {\Return F}
  }
  \uIf{state = on\-\_\-en\-trance\-\_\-curved\-\_\-path}{
    \uIf {$\|o-p\| \le s$}
    {
      Increment $j$, then let $o = o_{j} = (o_{x},o_{y})$ and $s = s_{j}$\; 
      state $\leftarrow on\-\_\-ex\-it\-\_\-curved\-\_\-path$\;
    }
    \uElse{
      $(F_{1},t) \leftarrow orbitPathFollowing (K_{TRVF}, c, r, p, \xi, w_{3}, k_{o}, K_{r}, v)$\;
      $F_{2} \leftarrow 1.5 K_{TRVF}\frac{o-p}{\|o - p\|}$\; 
      \Return $K_{TRVF}\frac{F_{1} + F_{2}}{\|F_{1} + F_{2}\|}$\;
    }
  }
  \uIf{state = on\-\_\-ex\-it\-\_\-curved\-\_\-path or state = on\-\_\-ex\-it\-\_\-straight\-\_\-path}{
    \uIf {$\|p-o_{j-1}\| > D$}{
      state $\leftarrow going\_to\_target$\;
      $F \leftarrow K_{TRVF}\frac{o-p}{\|o - p\|}$\;
      $F_{R} \leftarrow$ repulsiveForceFromTarget($o_{j-1},D$)\label{line:repfun2}\; 
      \Return $K_{TRVF}\frac{F + F_{R}}{\|F + F_{R}\|}$\; 
    }
    \uElse{
      \uIf{state = on\-\_\-ex\-it\-\_\-curved\-\_\-path}{
        $(F_{1},t) \leftarrow orbitPathFollowing (K_{TRVF}, c, r, p, \xi, w_{3}, k_{o}, K_{r}, v)$\; 
        \uIf {$t \le 0$}{
          $F_{2} \leftarrow 1.5 K_{TRVF}\frac{w_{3}-p}{\|w_{3} - p\|}$\;
          \Return $K_{TRVF}\frac{F_{1} + F_{2}}{\|F_{1} + F_{2}\|}$\;
        }
        \lElse{state $\leftarrow on\-\_\-ex\-it\-\_\-straight\-\_\-path$}
      }
      \uIf{state = on\-\_\-ex\-it\-\_\-straight\-\_\-path}{
        $(F,t) \leftarrow straightPathFollowing (K_{TRVF}, p, \xi, w_{3}, w_{4}, I_{d}, k_{s}, K_{r}, v)$\;
        \Return F\;
      }
    }
  }
  \caption{TRVF algorithm (continuation).}
  \label{alg:trvfcont}
\end{algorithm}

\section{Experiments and Results}
\label{sec:experimentresults}

To evaluate our approach, we executed several simulations using the Stage robot simulator \citep{PlayerStage}. We tested the algorithms for holonomic and non-holonomic robots in simulations. The robots measure 0.44 $\times$ 0.44 $\times$ 0.44 m$^{3}$. When testing non-holonomic robots, we used the control equations given in \citet{Luca94localincremental}. Our implementation considered the following equation for the repulsive forces \citep{Siegwart2004Introduction}:

\begin{equation}
{F} = 
\left\{
\begin{array}{ll}
-K_{Rep}\left(\dfrac{1}{d} - \dfrac{1}{I}\right)\dfrac{{q}-{p}}{d^3}, &  \mbox{if } d < I,\\
0, &  \mbox{otherwise,}
\end{array}
\right.
\label{eq:Forca}
\end{equation}
where $K_{Rep} > 0$ is a constant, $p$ the current position of the robot, $q$ the position of the neighbour, $d = \|q - p\|$ the Euclidean distance between $q$ and $p$, and $I$ the influence
radius.

We ran scenarios where robots are initially in a random position with a distance between 13 and 21 m from the target centre. After reaching the common target, robots will go towards the next one, which will be either to the left or to the right of the common target. This is decided randomly, according to a uniform probability (so roughly half of the robots would go to the left and half to the right). The new targets are aligned with the common target but far away in the $x$-axis.

Two kinds of experiments were performed, one for testing the SQF and TRVF algorithms and another to compare them with the state-of-the-art algorithms. 
Hyperlinks to the video of executions are available in the captions of each corresponding figure. They are in real-time so that the reader can compare the time and screenshots presented in the figures in this section with those in the supplied videos.\footnote{The source codes of each experimented algorithm are in \url{https://github.com/yuri-tavares/swarm-common-target-area-congestion}.}

\subsection{Proposed algorithms} 
\label{sec:experimentresultsalgs}

We compare here our algorithms with state-of-the-art algorithms for collision avoidance in the common target problem \citep{Marcolino2016}: PCC, EE, and PCC-EE. The default parameters used for the algorithms are shown in Table \ref{table:simulation_default}. For each experiment, we ran 40 executions, and we report in the graphs the average and the confidence interval such that $\rho = 0.01$. An appropriate statistical test with such $\rho$ is performed each time we mention statistical significance or say ``significantly better''. Since robots may not reach the target in some simulations, they were stopped after 60 minutes. This value was chosen because the total simulation time obtained at the runs where all robots reached the target without deadlocks was less than 45 minutes.

\begin{table}[t!]
  \centering
  \begin{tabular}{m{0.47\textwidth}m{0.11\textwidth}m{0.09\textwidth}m{0.2\textwidth}}
    \textbf{Parameter} & \textbf{Notation}  & \textbf{Value} & \textbf{Used by algorithm(s)}\\
    \hline
    Circular target area radius & $s$ & 3 m & All\\
    \hline
    Working radius of the algorithm around the target & $D$ & 13 m & All\\
    \hline
    Coefficient for repulsive forces & $K_{Rep}$ & 0.5 & All\\
    \hline
    Default influence radius & $I_{d}$ & 3.0 m& All\\
    \hline
    Minimum influence radius & $I_{min}$ & 1.0 m& SQF\\
    \hline
    SQF force magnitude & $K_{SQF}$ & 2.5 & SQF\\
    \hline
    Constant for proportional angular speed controller & $K_{r}$ & 3 & TRVF\\
    \hline
    TRVF force magnitude & $K_{TRVF}$ & 2.5 & TRVF\\
    \hline
    Maximum linear speed & $v$ & 1 m/s& TRVF\\
    \hline
    Constant for exponentiation in the straight line following & $k_{s}$& 1.1 & TRVF\\
    \hline 
    Constant for exponentiation in the orbit following vector fields & $k_{o}$ & 1.1 & TRVF\\
    \hline
    Angle of entry region & -- & $\frac{2\pi}{3}$  & EE, PCC-EE\\
    \hline
    Communication radius & -- & 3 m & PCC, PCC-EE\\
    \hline
    Radius of free region & -- & 3.7 m & PCC, PCC-EE\\
    \hline
    Angle of $\alpha$-area for waiting robot & -- & $\frac{23\pi}{36}$  & PCC, PCC-EE\\
    \hline
    Angle of $\alpha$-area for locked robot & -- & $\frac{\pi}{4}$  & PCC, PCC-EE\\
    \hline 
    Radius of $\alpha$-area & -- & 3 m &  PCC, PCC-EE\\ 
    \hline
    Number of cycles before sending a message & -- & 25 & PCC, PCC-EE\\
    \hline
    Number of cycles for testing if a waiting robot will change state & -- & 40 & PCC, PCC-EE\\
    \hline
    Radius of free region & -- & 3.7 m & PCC, PCC-EE\\
    \hline
    Radius of danger region & -- & 5.2 m & PCC, EE, PCC-EE\\
    \hline
  \end{tabular}
  \caption{Default values for simulation parameters with the notation used in this paper and algorithms that use it.}
  \label{table:simulation_default}
\end{table}

The next sections present the experiments and results for SQF and TRVF algorithms varying certain parameters, then a comparison between all algorithms is presented.

\subsubsection{SQF algorithm}

Figure \ref{fig:laneScreenshots} shows an execution of the SQF algorithm, with 100 non-holonomic robots with default parameters. Figure \ref{fig:laneScreenshots} (a) shows the initial positions of the robots around the target area (small circle in the centre). The larger circle around the target shows the distance $D$ where the rotational field will take place. Hence, robots inside the larger circle perform a rotational movement towards the corridor and soon reach the state in Figure \ref{fig:laneScreenshots} (b), where the first robots reach the corridor. These robots will directly move towards the target area (Figure \ref{fig:laneScreenshots} (c)), and then leave the region following a different rotational field, in yellow in Figures \ref{fig:laneScreenshots} (d) and (e). Eventually, all robots can reach the target. Robots in black are at a distance greater than $D$ and are now going towards their next target,  randomly chosen between left and right sides. 

\begin{figure}[t]
\centering
\subfloat[0s: Initial positions.]{\includegraphics[width=0.42\linewidth]{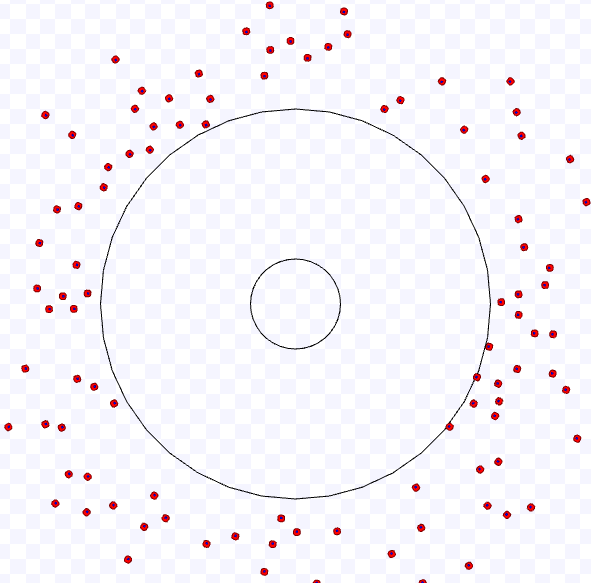}}
\hspace{1pt}
\subfloat[After 49.8 s.]{\includegraphics[width=0.42\linewidth]{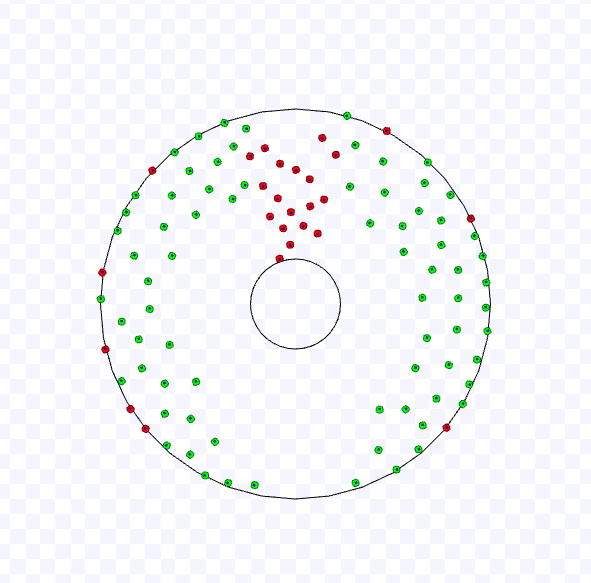}}
\\
\subfloat[After 142.8 s.]{\includegraphics[width=0.42\linewidth]{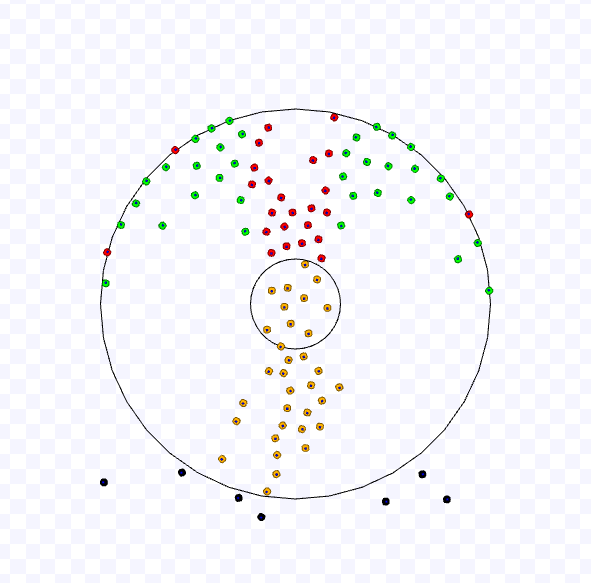}}
\hspace{1pt}
\subfloat[After 214.3 s.]{\includegraphics[width=0.42\linewidth]{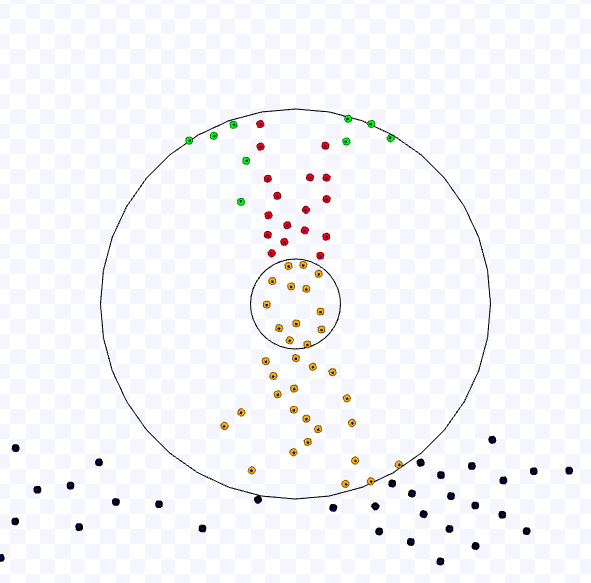}}
\caption{Screenshots of the SQF algorithm, with 100 non-holonomic robots with default parameters. The red, green and yellow robots are in state \emph{go\-ing\_to\_tar\-get}, \emph{go\-ing\_to\_cor\-ri\-dor} and \emph{leav\-ing\-\_\-tar\-get}, respectively. Black robots are going to their next target. This continues in Figure \ref{fig:laneScreenshots2}. Available on \url{https://youtu.be/3-d7Y7eViW4}.}
\label{fig:laneScreenshots}
\end{figure}

\begin{figure}[t]
\centering 
\addtocounter{subfigure}{4} 
\subfloat[After 285.7 s.]{\includegraphics[width=0.42\linewidth]{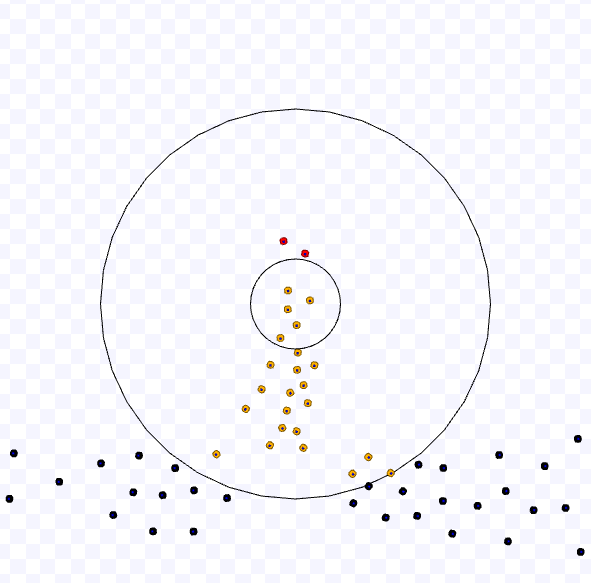}}
\hspace{1pt}
\subfloat[357.2 s: ending of the simulation.]{\includegraphics[width=0.42\linewidth]{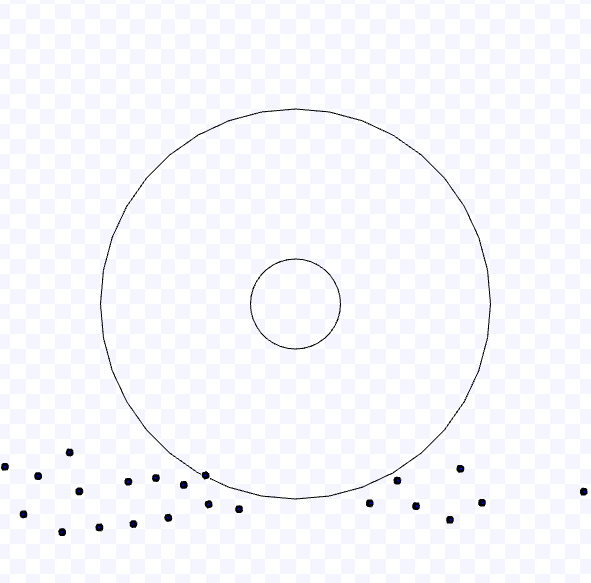}}
\caption{Continuation of Figure \ref{fig:laneScreenshots}.}
\label{fig:laneScreenshots2}
\end{figure}

\begin{figure}[t]
\centering
\subfloat[Holonomic]{\includegraphics[width=0.49\columnwidth]{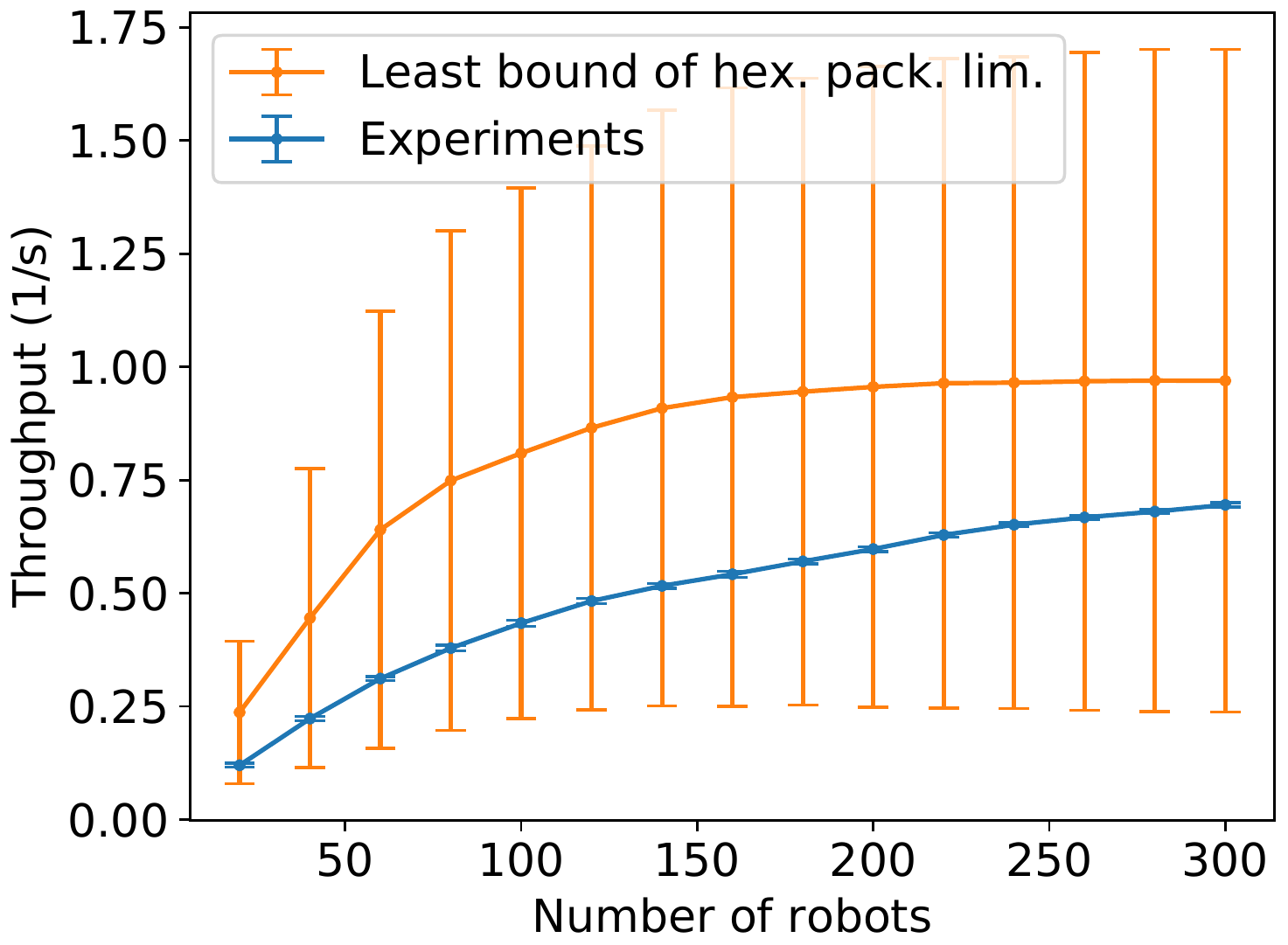}}
\subfloat[Non-holonomic]{\includegraphics[width=0.49\columnwidth]{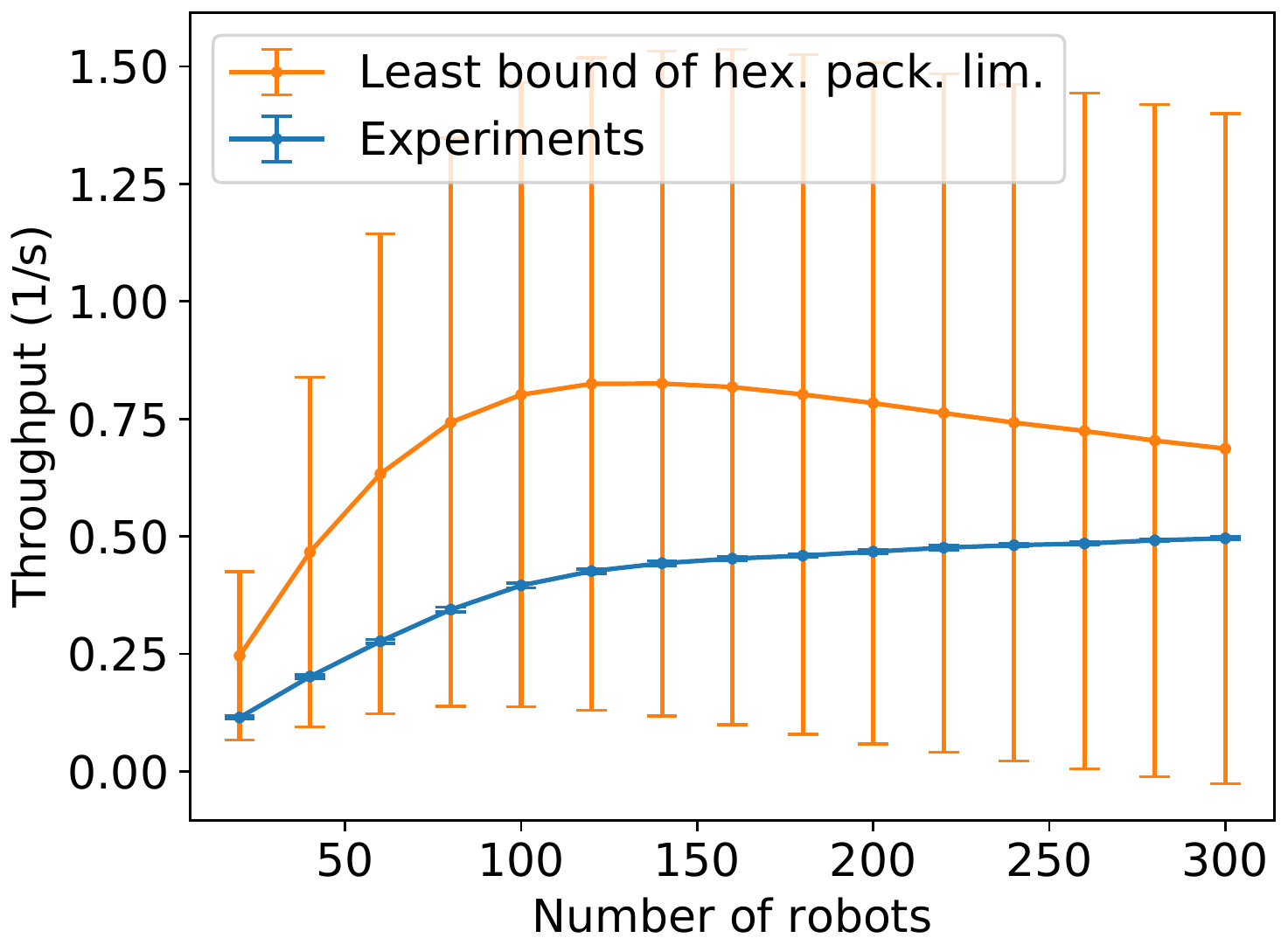}}
\caption{Throughput of SQF algorithm by the number of robots from 20 to 300 in steps of 20 for the experiments and the least bound (\ref{eq:exp:leastboundhex}) with $\theta = \pi/6$, using the mean distance between the robots and mean linear speed from experiments -- bars represent the shift of the means by one standard deviation to above and below.} 
\label{fig:r:nRobots}
\end{figure}

We analyse the throughput for a growing number of robots, considering holonomic and non-holonomic robots.  Figure \ref{fig:r:nRobots} displays the results for the experiments in comparison with the least bound of asymptotic throughput (that is, the limit of the throughput when the number of robots tends to infinity) from the hexagonal packing corridor strategy that inspired the SQF algorithm. This value was derived in our concurrent theoretical work \citep{arxivTheory} and is given by

\begin{equation} 
\frac{4vs}{\sqrt{3}d^{2}} - \frac{2 v  \cos(\theta -\pi/6)}{\sqrt{3}d}, 
\label{eq:exp:leastboundhex}
\end{equation}
for the linear speed $v$, the distance between each robot of $d$, an angle $\theta$ of the hexagonal packing arrangement and target region radius of $s$. This value is minimised by  $\theta = \pi/6$. 

In Figure \ref{fig:r:nRobots}, we are using it with the mean distance between each robot and its closest neighbour and mean linear speed in all experiments for each number of robots. These values do not follow a normal distribution. Thus, instead of plotting the confidence interval with $\rho = 0.01$, we used these means shifted to above and below by one standard deviation to calculate the interval shown in Figure \ref{fig:r:nRobots}. Observe in this figure that the results obtained from the experiments are still below the upper bound obtained by the mean values but inside the one standard deviation interval. Although the high variation in the distance between the robots and linear speed contributes to the difference, the SQF algorithm forms a corridor only in the upper part of the circle with a radius of $D$ around the target. Hence, the robots still need time to get to the corridor. Also, the positions of the robots in the corridor are not too compacted.
 
\subsubsection{TRVF algorithm}

For the repulsive force away from the previous target region we used one similar to (\ref{eq:Forca}) but the distance is considered from the robot position and the circle with radius $D$ centred at the target position, and the influence radius is $D$, that is, 

\begin{equation}
{F} = 
\left\{
\begin{array}{ll}
-K_{Rep}\left(\dfrac{1}{d} - \dfrac{1}{D}\right)\dfrac{1}{d^2}\dfrac{{o}-{p}}{\|{o}-{p}\|}, &  \mbox{if } d < D,\\
0, &  \mbox{otherwise,}
\end{array}
\right.
\label{eq:Forca2}
\end{equation}
where $K_{Rep} > 0$ is a constant, ${p}$ the current position of the robot, ${o}$ the target centre position, and $d = \|{o} - {p}\| - D$ is the distance from the robot position to the circle.

Figure \ref{fig:TRVFScreenshots} displays an execution of the TRVF algorithm, with 100 non-holonomic robots and four lanes with default parameters. Figure \ref{fig:TRVFScreenshots} (a) shows the initial positions of the robots around the target area (small circle in the centre). The larger circle around the target shows the distance $D$, and the robots will avoid it when they go to the next target using (\ref{eq:Forca2}). Each robot goes to the entering lane next to its position and follows the force field (Figure \ref{fig:TRVFScreenshots} (b)). The robots avoid crossing the target region and follow the border of the circle with radius $D$ when going to the next target in Figures \ref{fig:TRVFScreenshots} (c), (d) and its continuation on Figure \ref{fig:TRVFScreenshots2} so that all robots can reach the target. With this algorithm, the cluttering is distributed through the entrances and the target region according to the number of lanes, while the leaving lanes become free. However, without the TRVF algorithm, the clutter would concentrate only next to the target region.

\begin{figure}[t!]
  \centering
  \subfloat[0s: Initial positions.]{\includegraphics[width=0.4\linewidth]{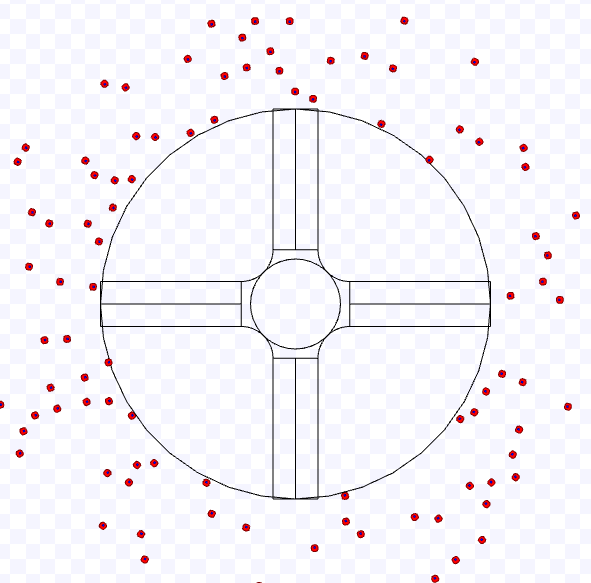}}
  \hspace{1pt}
  \subfloat[After 89.5 s.]{\includegraphics[width=0.4\linewidth]{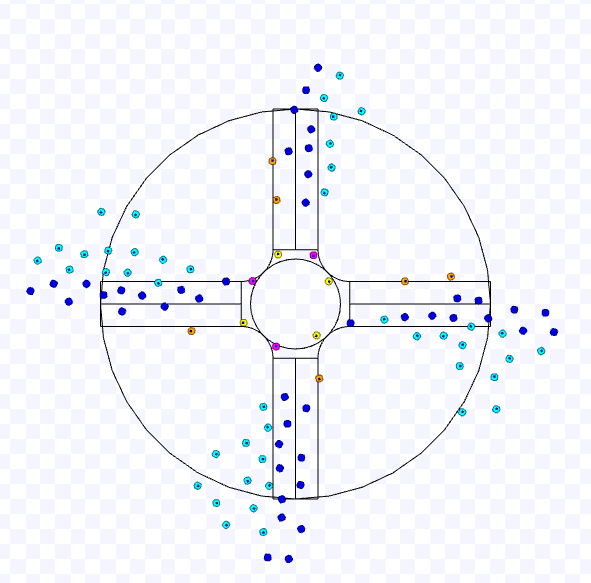}}
  \\
  \subfloat[After 179 s.]{\includegraphics[width=0.4\linewidth]{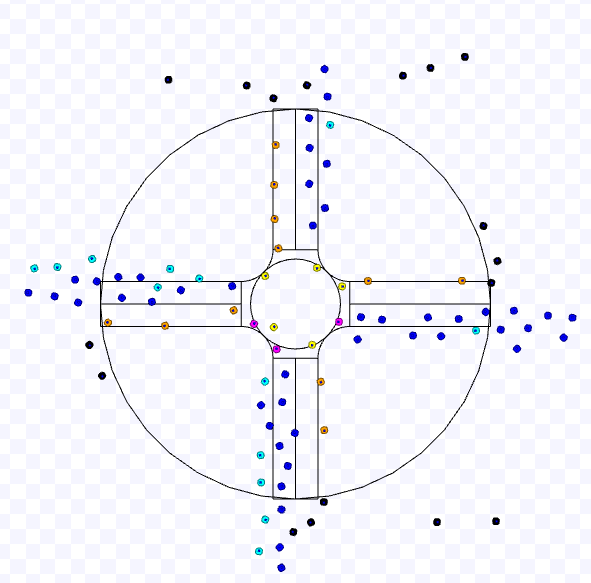}}
  \hspace{1pt}
  \subfloat[After 268.5 s.]{\includegraphics[width=0.4\linewidth]{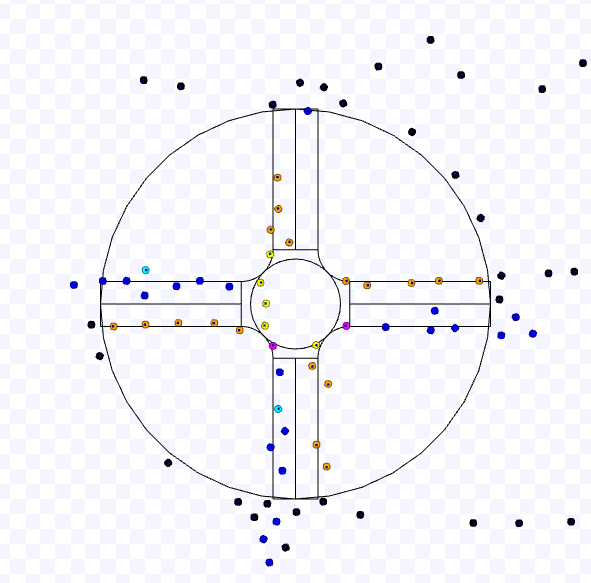}}
  \caption{Screenshots of the TRVF algorithm, for 100 non-holonomic robots, four lanes and default parameters. The colour of the robots are the same for their states shown in Section \ref{sec:proposedalgs}, that is, red, light blue, blue, magenta, yellow and orange for   \emph{go\-ing\-\_\-to\-\_tar\-get},   \emph{go\-ing\-\_\-to\-\_\-en\-trance\-\_\-straight\-\_\-path},   \emph{on\-\_\-en\-trance\-\_\-straight\-\_\-path},   \emph{on\-\_\-en\-trance\-\_\-curved\-\_\-path},   \emph{on\-\_\-ex\-it\-\_\-curved\-\_\-path}  and  \emph{on\-\_\-ex\-it\-\_\-straight\-\_\-path}  states, respectively. Black robots are going to their next target. This continues in Figure \ref{fig:TRVFScreenshots2}.  Available on \url{https://youtu.be/MRzXS_9I2Ls}.}
  \label{fig:TRVFScreenshots}
\end{figure}

\begin{figure}[t!]
\centering 
\addtocounter{subfigure}{4} 
  \subfloat[After 358 s.]{\includegraphics[width=0.4\linewidth]{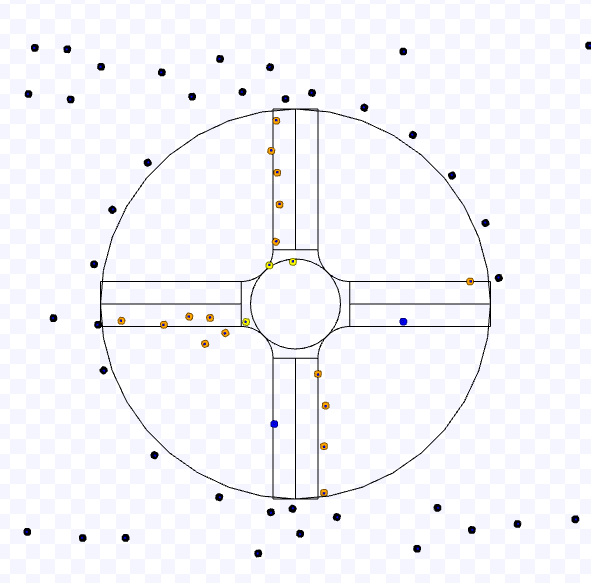}}
  \hspace{1pt}
  \subfloat[447.5 s: ending of the simulation.]{\includegraphics[width=0.4\linewidth]{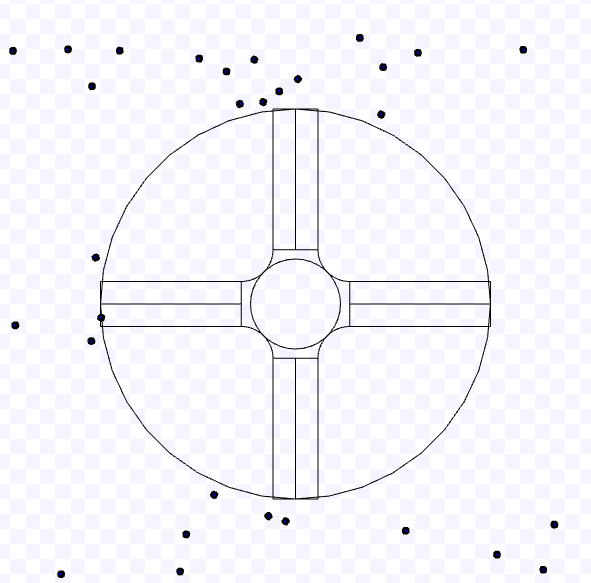}}
\caption{Continuation of Figure \ref{fig:TRVFScreenshots}.}
\label{fig:TRVFScreenshots2}
\end{figure}

\begin{figure}[t]
\centering
\subfloat[Holonomic]{\includegraphics[width=0.49\columnwidth]{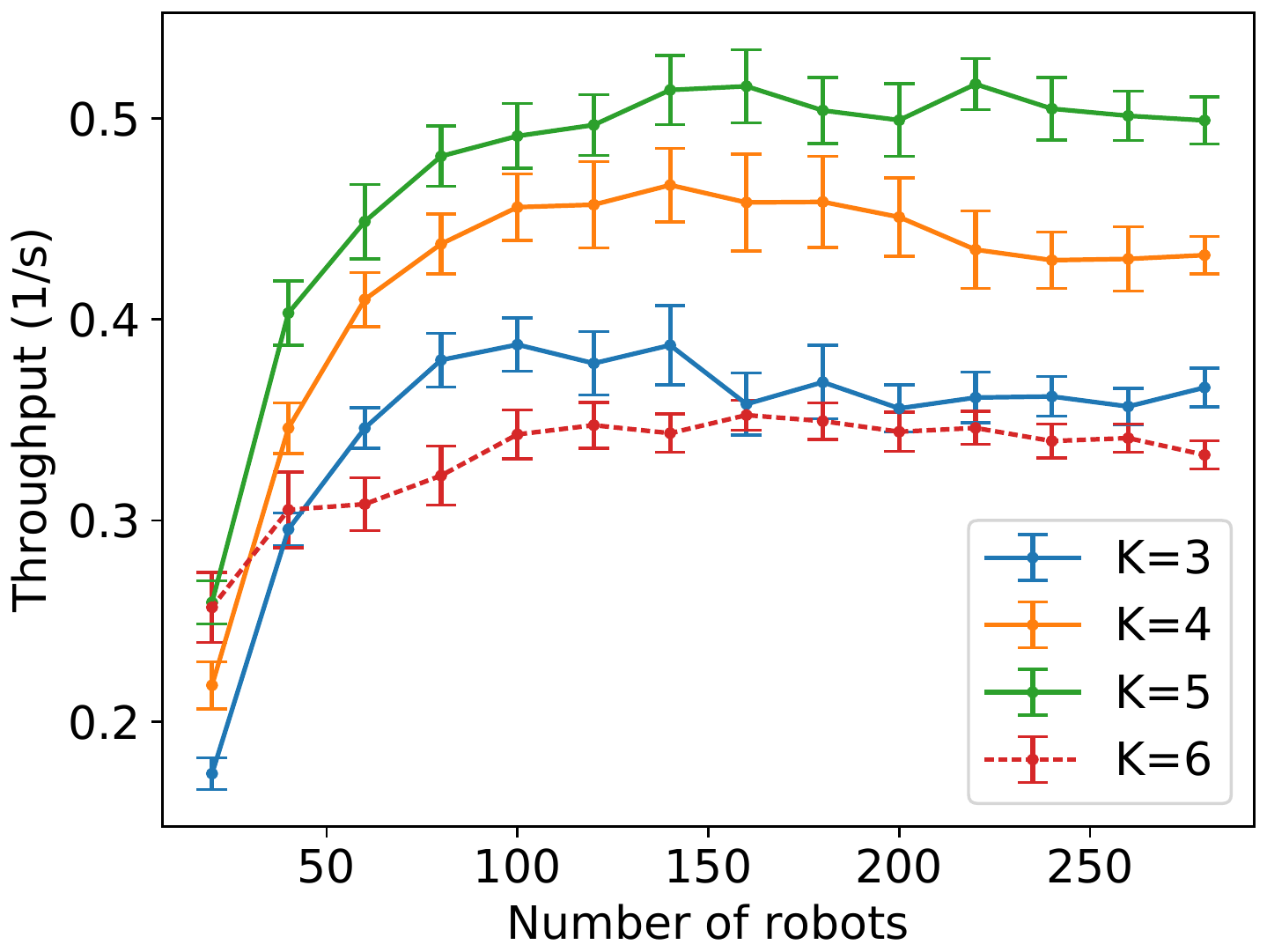}}
\subfloat[Non-holonomic]{\includegraphics[width=0.49\columnwidth]{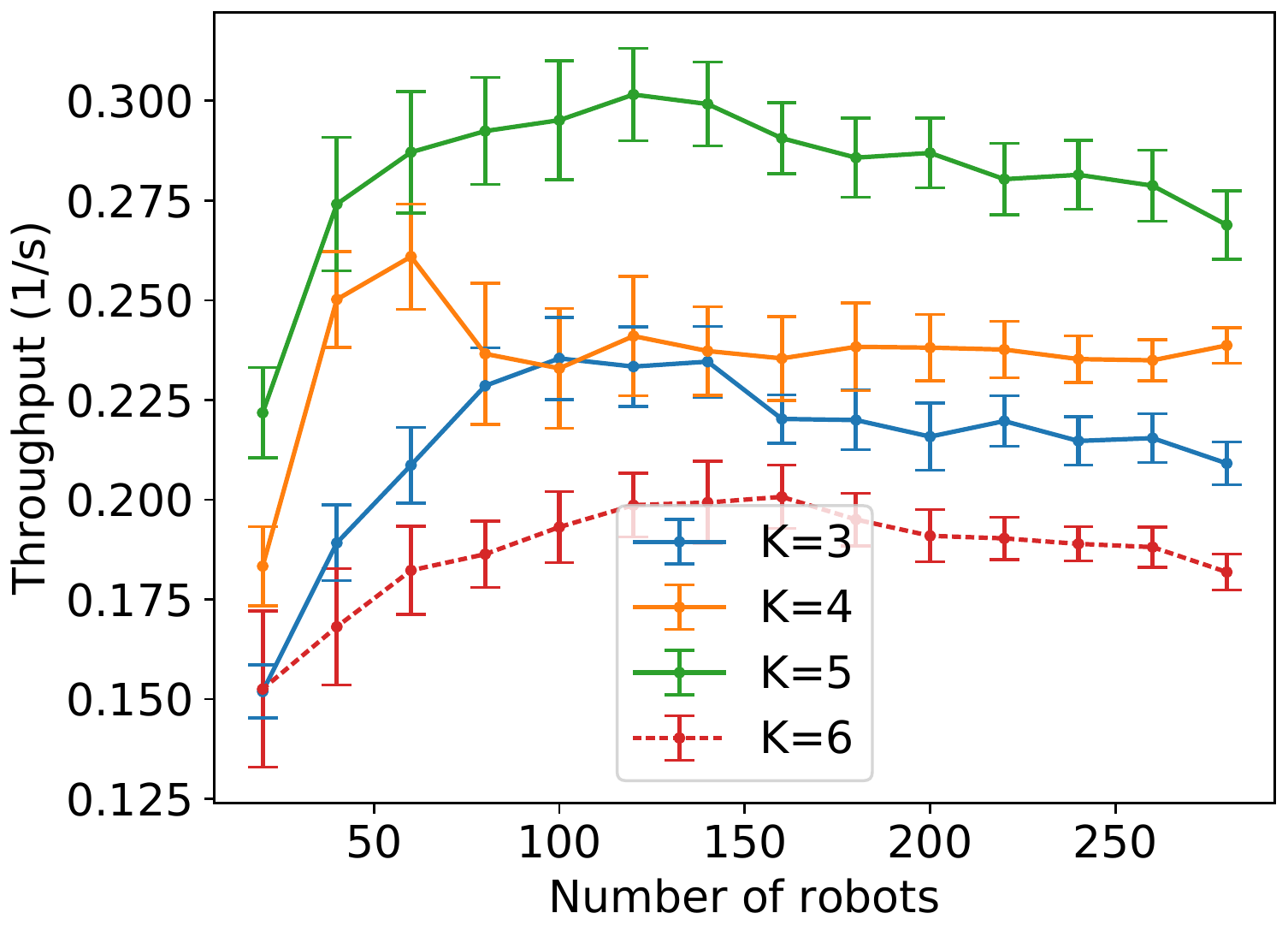}}
\caption{Throughput of TRVF algorithm by the number of robots from 20 to 300 in steps of 20 for different values of the number of lanes (K).}
\label{fig:KtestsTRVF}
\end{figure}

We performed simulations to compare different K values for the defaults parameters for a number of robots ranging from 20 to 300 with increments of 20 with the default values in Table \ref{table:simulation_default}. The results are in Figure \ref{fig:KtestsTRVF}. 

Firstly, we compare with the results for the touch and run strategy by calculating the asymptotic throughput for the allowed values of K -- that is, $K \in \{3,\dots,6\}$ -- following the bound found in\citep{arxivTheory}

\begin{equation} 
\begin{aligned} 
&\frac{Kv }{\max(d,d')} \text{ for } 
 d' &= 
    \begin{cases}
       r  (\pi - \alpha)  + \frac{d - 2  r  \cos(\alpha / 2)}  {\sin(\alpha / 2)}, & \text{ if } 2  r  \cos(\alpha / 2) < d,\\
      2  r  \arcsin\left( \frac{d}{2 r}\right), & \text{ otherwise. }
    \end{cases} 
\end{aligned}
\label{eq:exp:throughputhitandrunlimit}
\end{equation}

For K equals 3, 4, 5 and 6, the asymptotic throughput rounded to three decimal places are 0.994, 1.2, 1.099 and 1, respectively. Thus, $K = 4$ would be the best number of lanes for the touch and run strategy. However, as in the TRVF experiments, the distances between robots and linear speeds change over time, and the movement of the robots is influenced by the other robots in their neighbourhood. Therefore, the best K was 5 according to the results in Figure \ref{fig:KtestsTRVF}.

\begin{figure}[t!]
  \centering
  \subfloat[K = 3 and after 202.8 s.  Available on \url{https://youtu.be/0U6ajiBtYw8}.]{\includegraphics[width=0.49\linewidth]{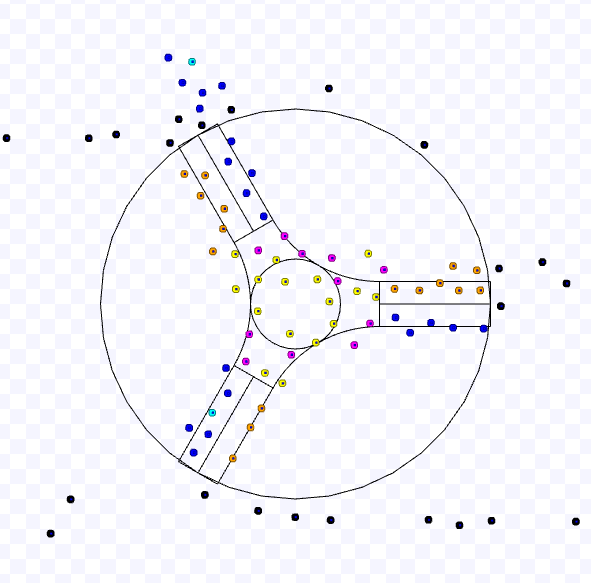}}
  \hspace{1pt}
  \subfloat[K = 4 and after 155.5 s. Available on \url{https://youtu.be/D-_2VK5JRYg}.]{\includegraphics[width=0.49\linewidth]{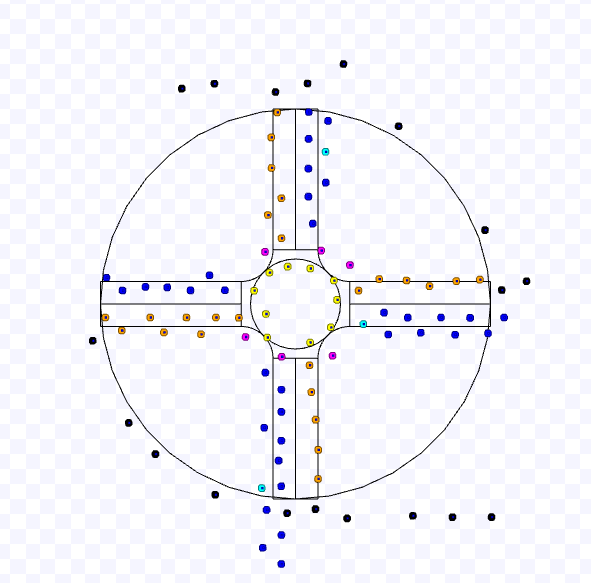}}
  \\
  \subfloat[K = 5 and after 156.1 s. Available on \url{https://youtu.be/z92SNJ8ugHs}.]{\includegraphics[width=0.49\linewidth]{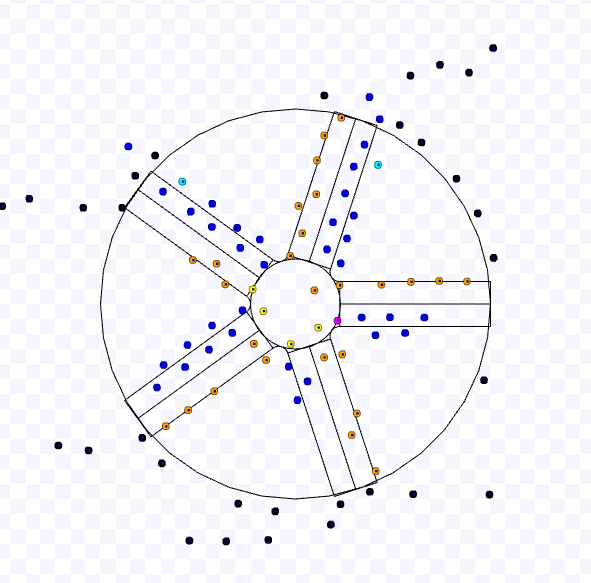}}
  \hspace{1pt}
  \subfloat[K = 6 and after 177.3 s. Available on \url{https://youtu.be/BYd8VncXnCQ}.]{\includegraphics[width=0.49\linewidth]{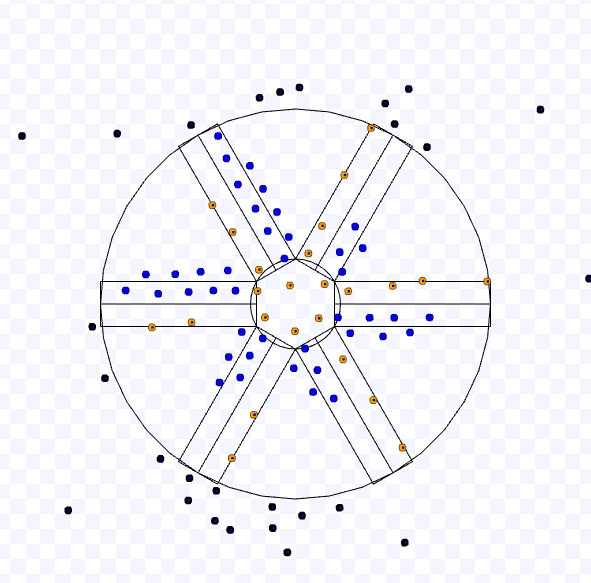}}
  \caption{Screenshots of the TRVF algorithm, for 100 holonomic robots, default parameters and different numbers of lanes (K) at the middle of the execution.}
  \label{fig:differentKexecs}
\end{figure}

Figure \ref{fig:differentKexecs} shows a screenshot in the middle of the execution for the allowed K values and 100 holonomic robots. Using $K=6$, the default values $s = 3$ m and $d = I_{d} = 3$ m in (\ref{eq:relationangles}) gives $r=0$, so the TRVF algorithm will run but no curved trajectory is made. In this case, the entrance and exiting straight lanes lie exactly in the target region. This could be a shorter trajectory, as the robots do not need to make the curved turn, but congestions happen in the target region (Figure \ref{fig:differentKexecs} (d)). In fact, for any K displayed in Figure \ref{fig:differentKexecs}, observe that some robots are trapped inside the target area. They were pushed inside the area due to repulsive forces from other robots, and they inflicted those forces on the robots getting near the target region from the curved trajectory, consequently causing congestion as well. However, for the best $K=5$, congestion is minimised.

\begin{figure}[t]
\centering
\subfloat[Holonomic.]{\includegraphics[width=0.49\columnwidth]{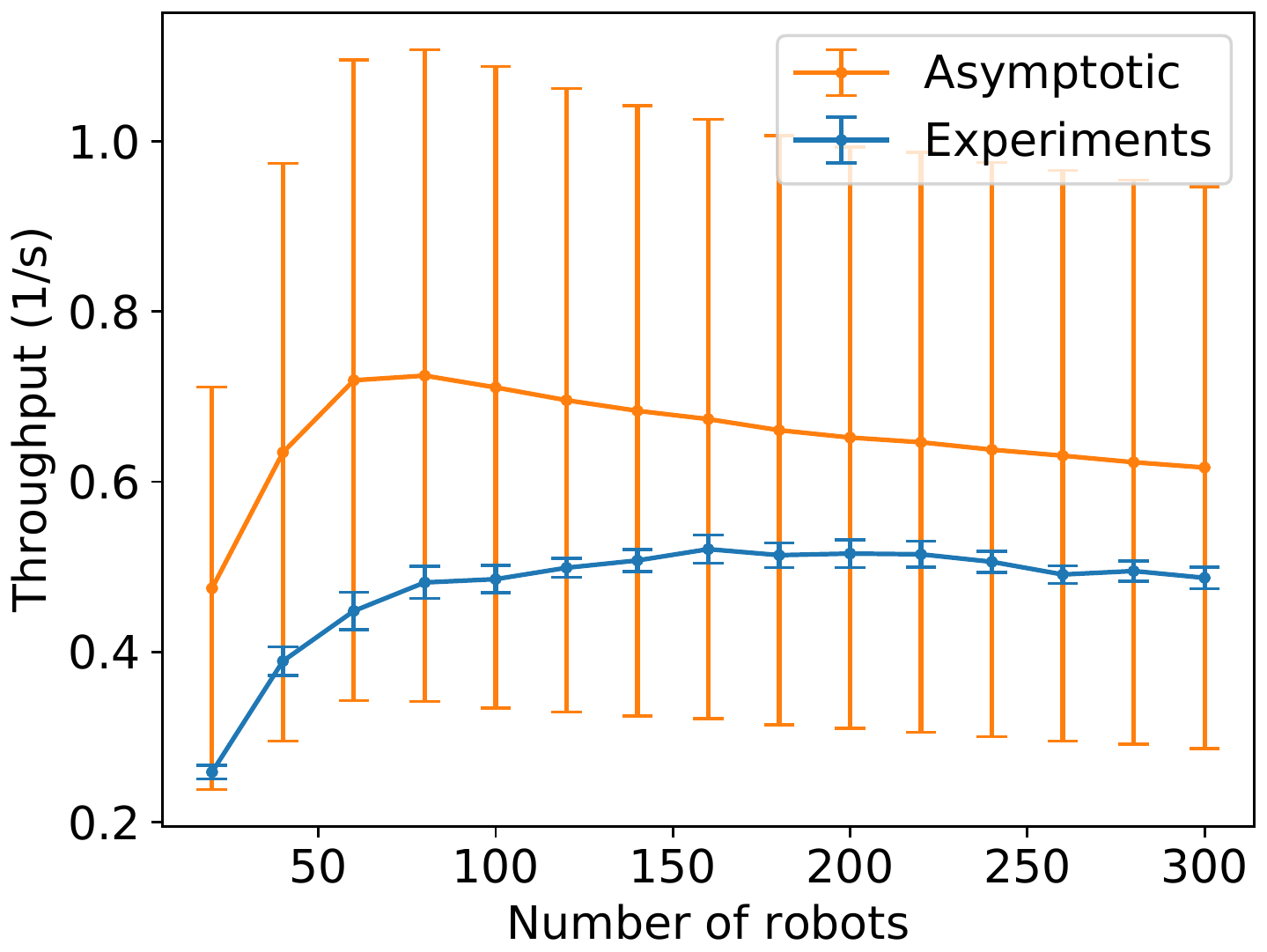}}
\subfloat[Non-holonomic.]{\includegraphics[width=0.49\columnwidth]{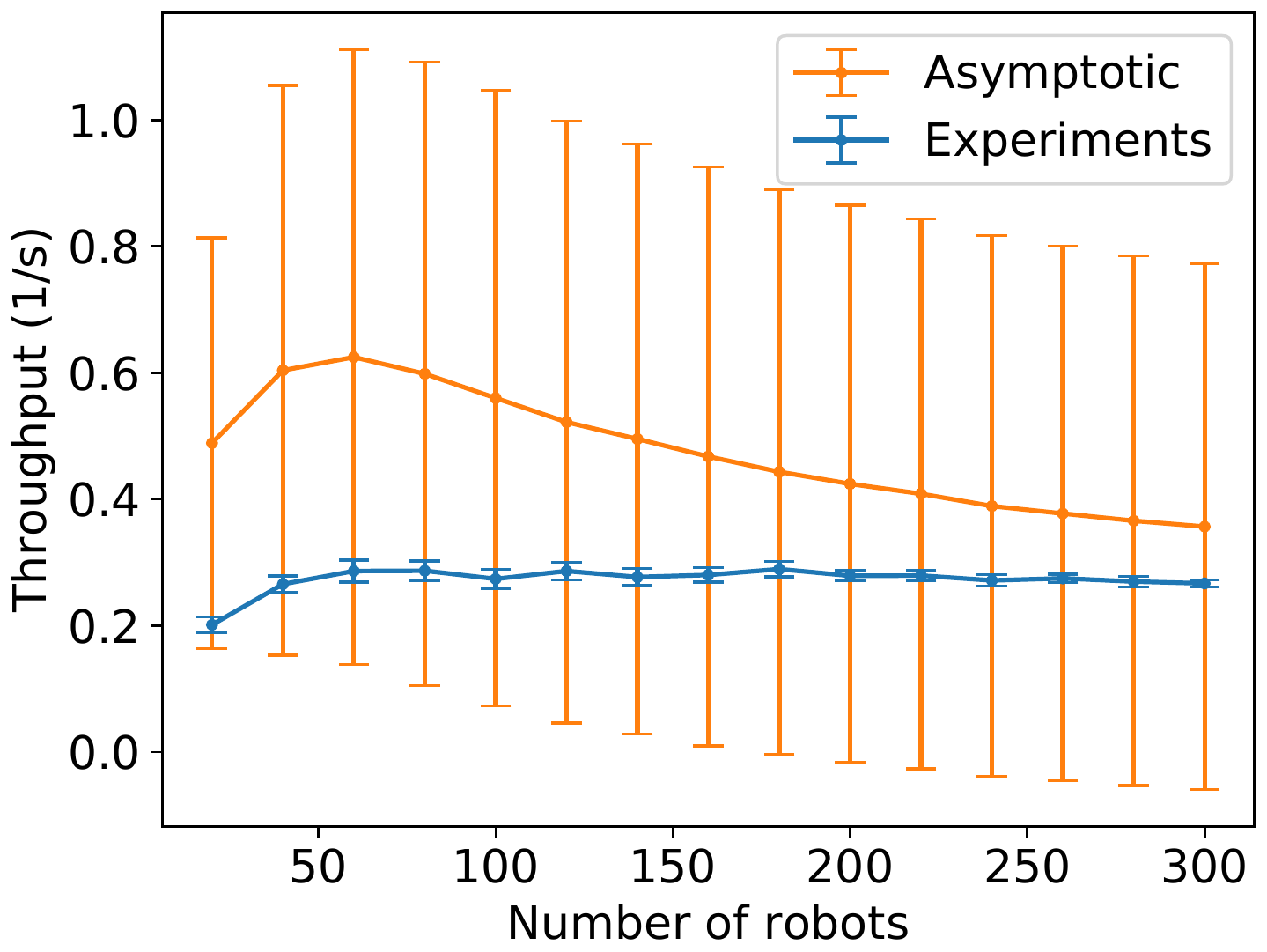}}
\caption{Throughput of TRVF algorithm by the number of robots from 20 to 300 in steps of 20 for the experiments, and asymptotic throughput using $K=5$, the mean distance between the robots and mean linear speed from experiments -- bars represent the shift of the means by one standard deviation to above and below.}
\label{fig:T:nRobots}
\end{figure}

As done for the SQF algorithm, we analyse the throughput for a growing number of robots, considering holonomic and non-holonomic robots. Figure \ref{fig:T:nRobots} displays the results for the experiments in comparison with the theoretical maximum throughput value. The theoretical maximum throughput was obtained by (\ref{eq:exp:throughputhitandrunlimit}) for $K=5$ with the mean distance between the robots and mean speed in all experiments for each number of robots. As before, in order to plot the offset from the mean values, we used the standard deviation instead of the confidence interval here because these values do not follow a normal distribution. From that figure, the experimental throughput is still below the upper bound obtained by the mean values but inside the one standard deviation interval. The difference between the experimental data and the theoretical value obtained from the mean values occurs because, by using variable linear speed, the robots are not constantly paced towards the target. Additionally, next to the curve, the robots deviate from the trajectory to avoid bumping into each other.

\subsection{Comparison with state-of-art algorithms}

In this section, we compared the algorithms regarding (i) the throughput, (ii) the time the swarm reaches the target area, (iii) the time for leaving it averaged by the number of robots, and (iv) the total simulation time. We only consider the runs that succeed in terminating within 60 minutes in Section \ref{sec:exp:constantradius} and 20 minutes in Section \ref{sec:exp:varyingradius}. This termination time is lower in the later section because we use 100 robots, while in the former, up to 300. We observed that the robots running algorithms PCC, EE and PCC-EE for the experiments of Section \ref{sec:exp:varyingradius} entered deadlock in executions longer than 20 minutes for 100 robots.

The total simulation time is obtained when the last robot in the experiment arrives at the outer circle with a radius of $D$ after everyone has reached the target area. The reaching time is measured from the last robot in the swarm that gets to the target area. The leaving time is taken from the moment a robot reaches the target area until it arrives at the outer circle. As done in \cite{Marcolino2016}, to measure the effectiveness of the algorithms on the crowd reduction for leaving the target area, we summed the leaving time of every robot and divided it by the number of robots in the experiment.

Additionally, we performed tests with the target area with constant radius (as in Table \ref{table:simulation_default}) in Section \ref{sec:exp:constantradius} and varying small values in Section \ref{sec:exp:varyingradius}.

\subsubsection{Comparison for constant target size}
\label{sec:exp:constantradius}

\paragraph{Comparison of the reaching time and throughput}

\begin{figure}[t]
\centering
\subfloat[Holonomic]{\includegraphics[width=0.49\columnwidth]{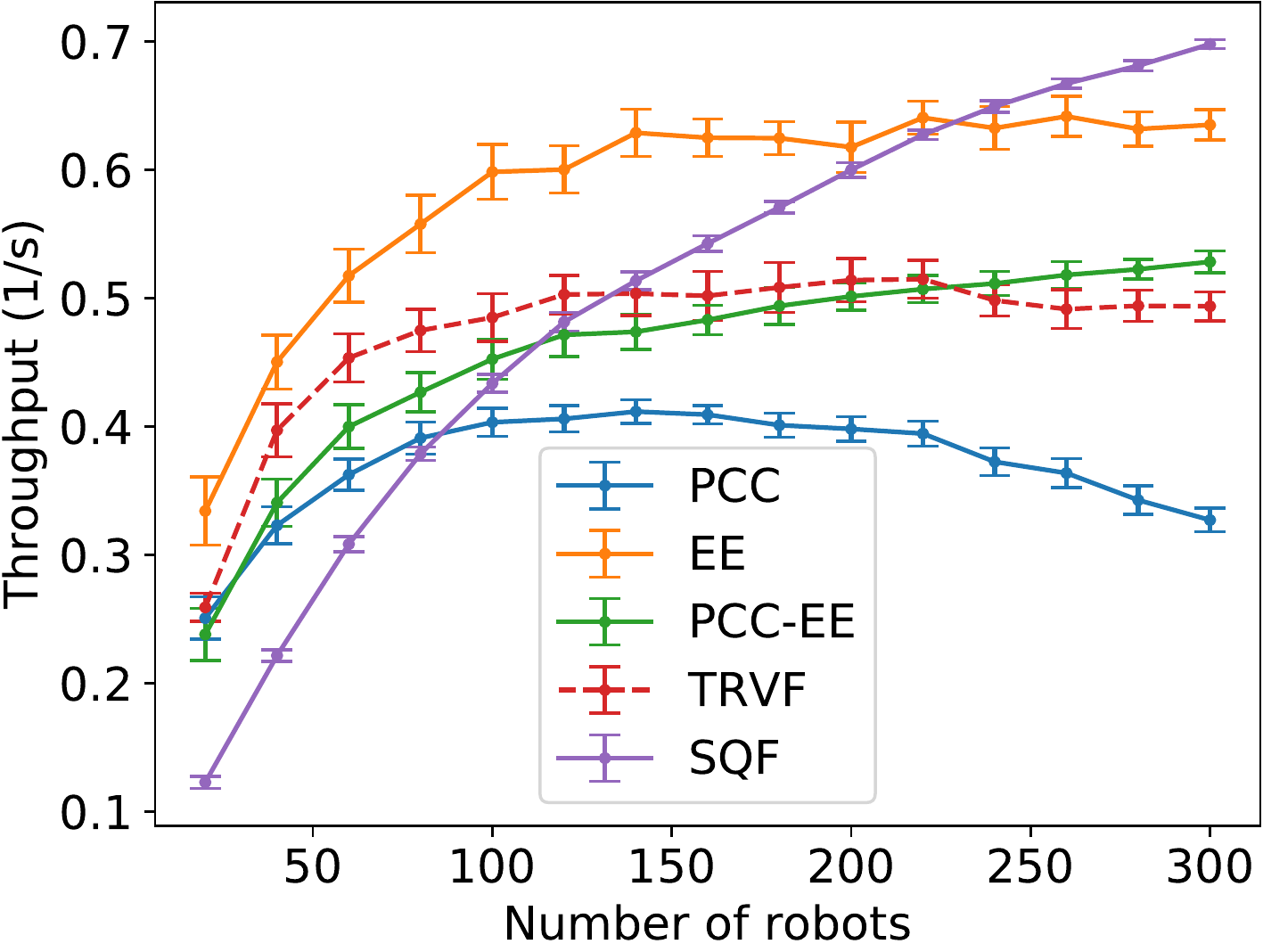}}
\subfloat[Non-holonomic]{\includegraphics[width=0.49\columnwidth]{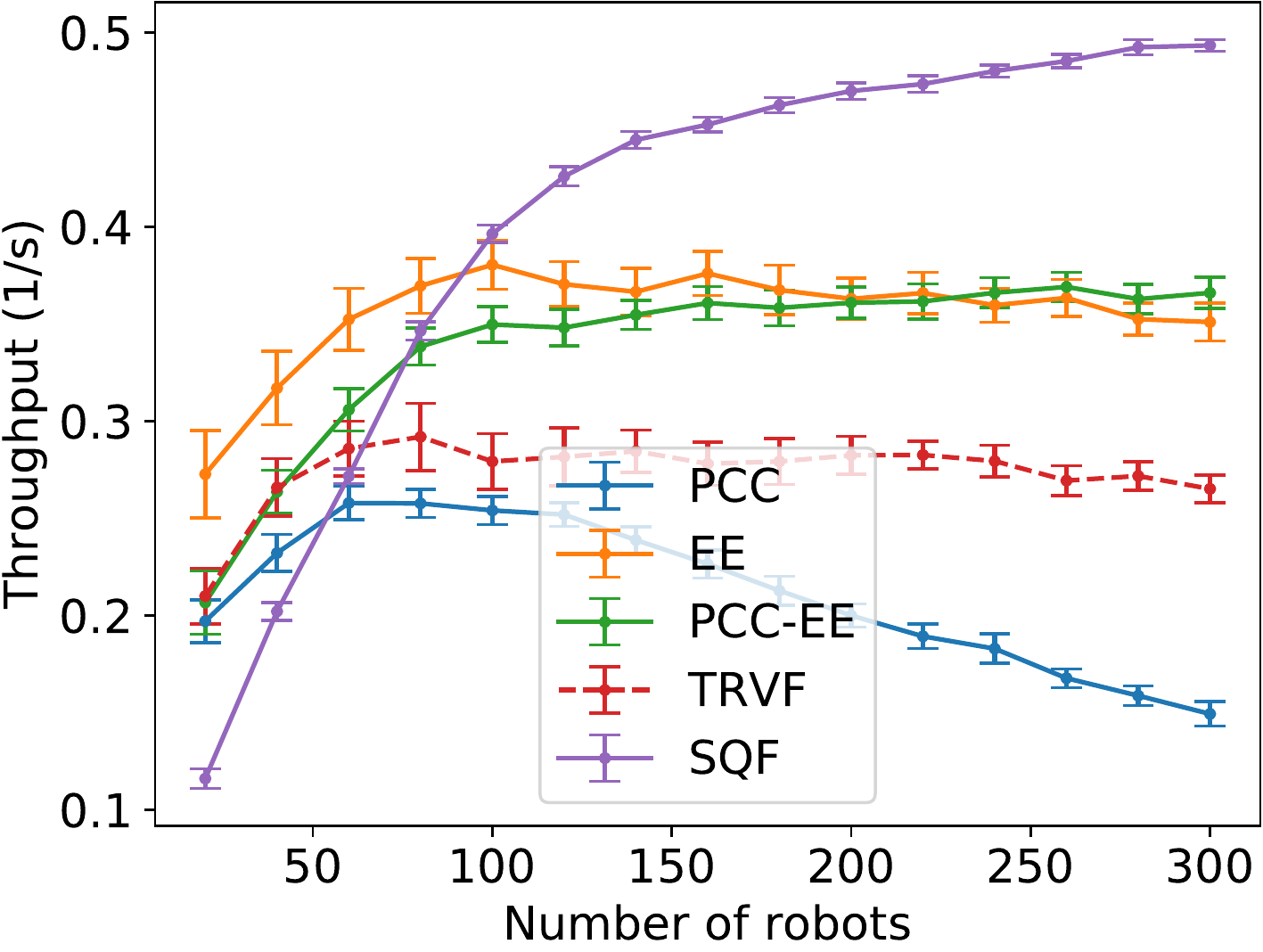}}
\caption{Throughput comparison of the algorithms for a number of robots from 20 to 300 in steps of 20.}
\label{fig:algsTh}
\end{figure}

\begin{figure}[t]
\centering
\subfloat[Holonomic]{\includegraphics[width=0.485\columnwidth]{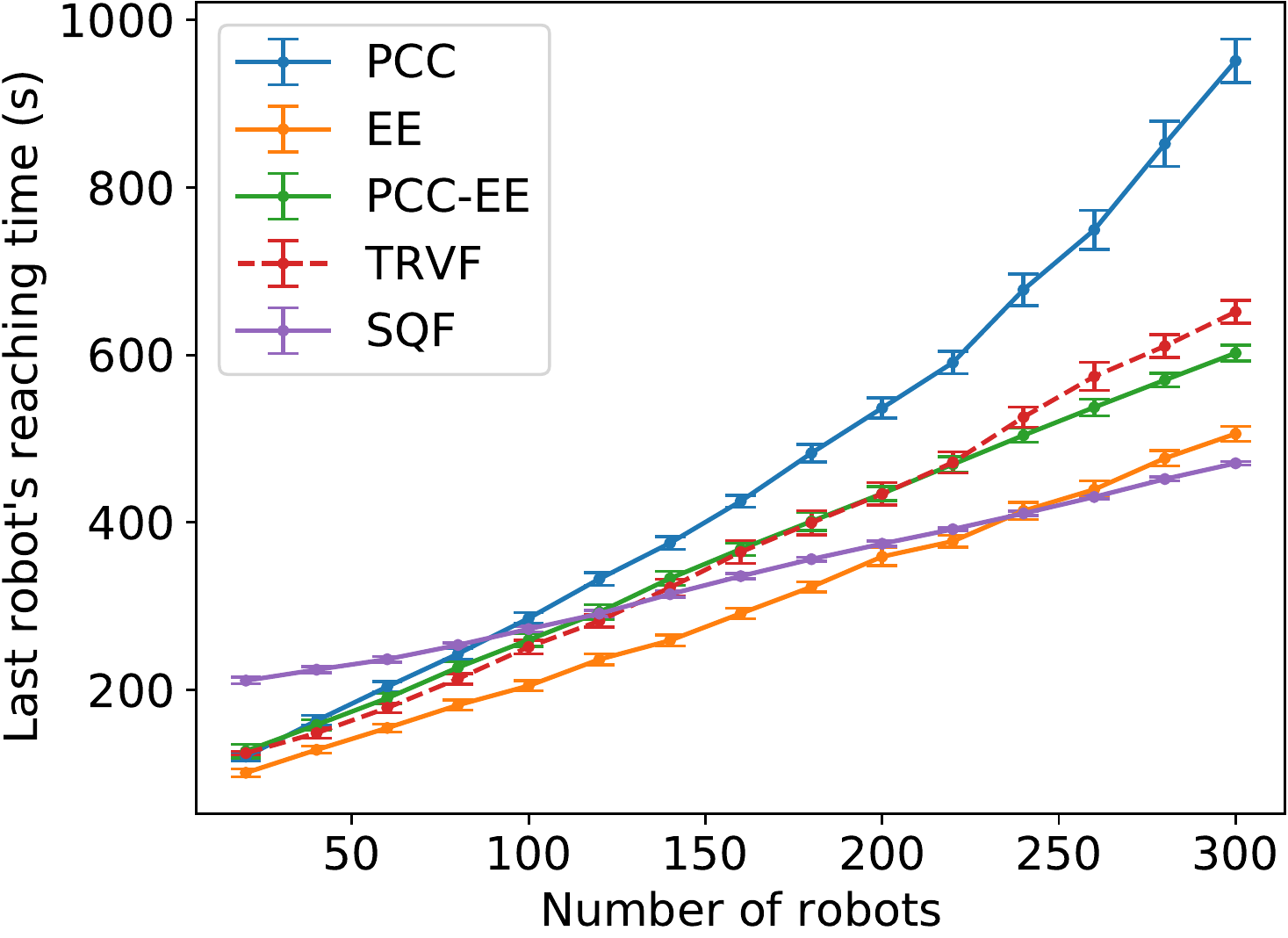}}
\subfloat[Non-holonomic]{\includegraphics[width=0.495\columnwidth]{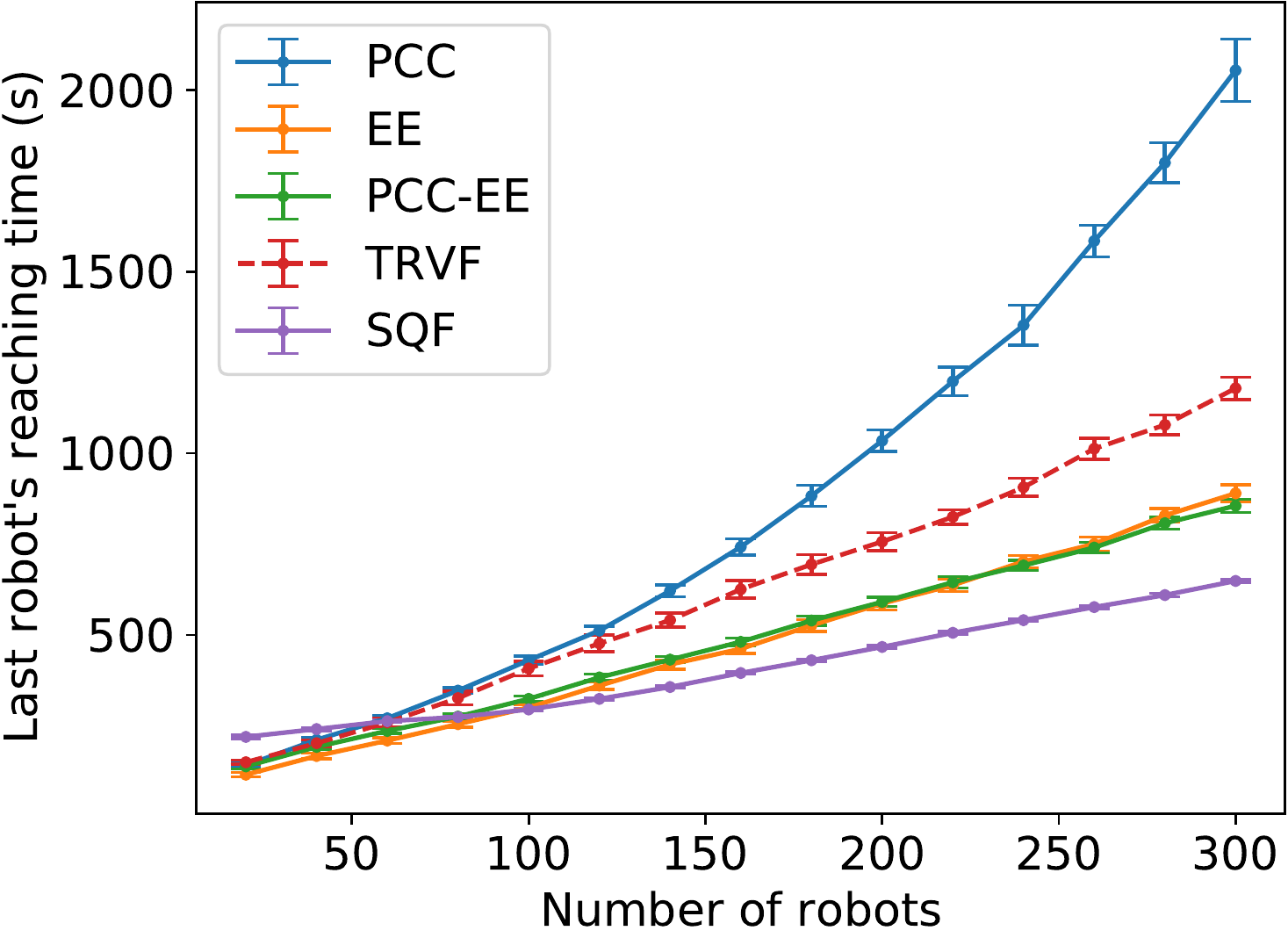}}
\caption{Comparison of the time to reach the target of the algorithms for a number of robots from 20 to 300 in steps of 20.}
\label{fig:algsTime}
\end{figure}

Figures \ref{fig:algsTh} and \ref{fig:algsTime} show the comparison for a varying number of robots of the target area throughput and the reaching time, respectively. For both types of robots, throughput and time increase with the number of robots. Additionally, observe in the comparisons that higher throughput reflects a lower arrival time. 

Moreover, we notice in the throughput graph that, for holonomic robots from 240 robots, the SQF algorithm is significantly better than all other algorithms and, from 100 robots, for non-holonomic ones. In the reaching time graph, this only occurs from 280 for holonomic and 120 for non-holonomic ones with statistical significance by t-test. However, below these values, EE is still better by throughput. Excluding the EE algorithm, the TRVF algorithm is better than all the remaining algorithms from 40 to 120 individuals for holonomic robots but, for non-holonomic robots, it is only better than SQF and PCC algorithms just for 40 and 60 robots. 

\begin{figure}[t]
\centering
\subfloat[Holonomic]{\includegraphics[width=0.49\columnwidth]{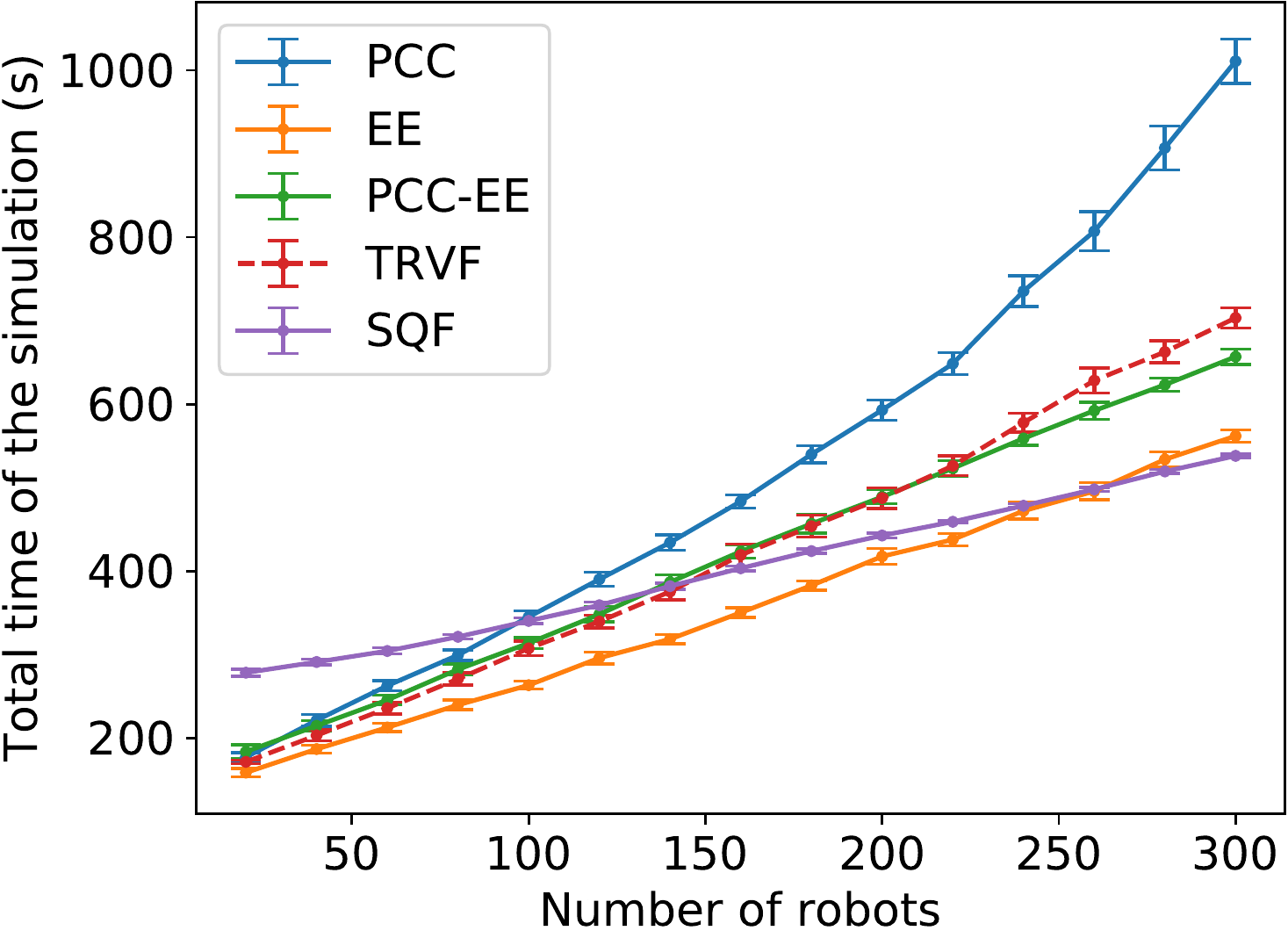}}
\subfloat[Non-holonomic]{\includegraphics[width=0.49\columnwidth]{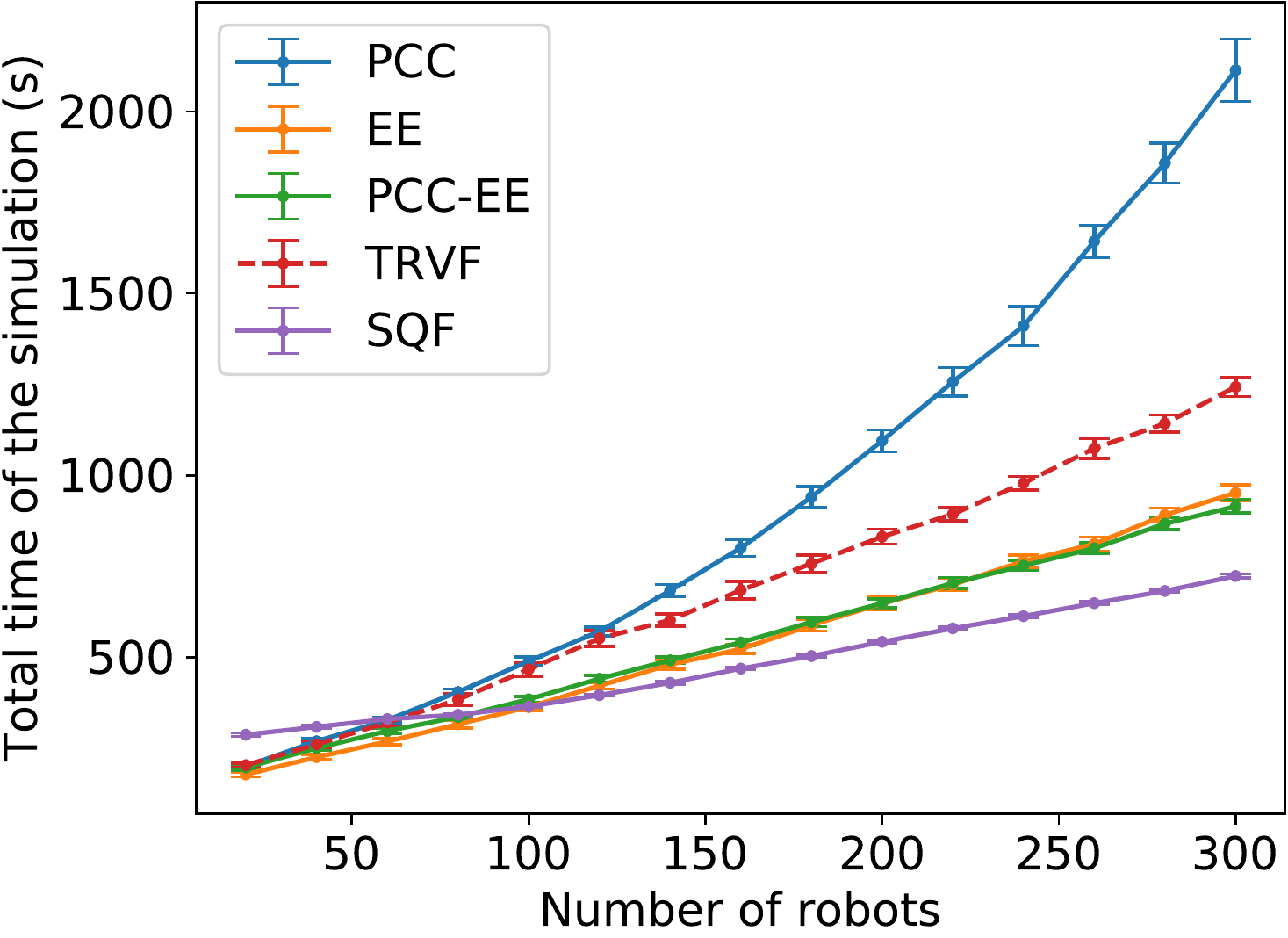}}
\caption{Total simulation time comparison of the algorithms for a number of robots from 20 to 300 in steps of 20.}
\label{fig:algsTotalTime}
\end{figure}

Furthermore, unpaired t-tests with $\rho = 0.01$ returned that the reaching time intervals of SQF and EE algorithms for 240 and 260 holonomic robots -- Figure \ref{fig:algsTime} (a)) -- have the same mean. These tests showed that the reaching time intervals for 100 non-holonomic robots of SQF and EE algorithms -- Figure \ref{fig:algsTotalTime} (b) -- also have the same mean.

Comparing only TRVF and SQF by the target reaching time, the former is better than the latter until 120 individuals for holonomic robots and up to 40 for the non-holonomic case with statistical significance. Unpaired t-tests returned that SQF and TRVF reaching time mean intervals have different means except for 140 holonomic and 60 non-holonomic robots. As the SQF algorithm organises a queue above the target region, the robots must initially follow the way until reaching that queue. After that, the robots flow through it. On the other hand, in the TRVF, the robots go directly to the target region. However, as the number of robots increases, more congestions happen near this region because of the repulsive forces caused by the robots doing the curve.

Observe as well that, although SQF has higher throughput than TRVF from some number of robots, the comparison of their corresponding inspiration strategies, hexagonal packing and touch and run, had a different result with respect to the asymptotic throughput presented in \citep{arxivTheory}. For calculating asymptotic throughput, we considered constant the maximum linear speed and the minimum distance between robots. Using only asymptotic throughput, we can evaluate which one is the best strategy in a scenario with those fixed quantities. For robots using artificial fields, getting an explicit asymptotic throughput equation is difficult due to the changeability of the speed and the distance between the robots, but this does not prevent using experimental throughput for comparisons. Accordingly, we notice that this changeability and the effect of the other robots in the trajectory yield the SQF being better than TRVF, although the touch and run is better than hexagonal packing. Even so, the analytically calculated throughputs are still upper bounds on the ones obtained from simulations, as observed in Figures \ref{fig:r:nRobots} and \ref{fig:T:nRobots}.

\paragraph{Comparison of the average leaving time}

\begin{figure}[t]
\centering
\subfloat[Holonomic]{\includegraphics[width=0.49\columnwidth]{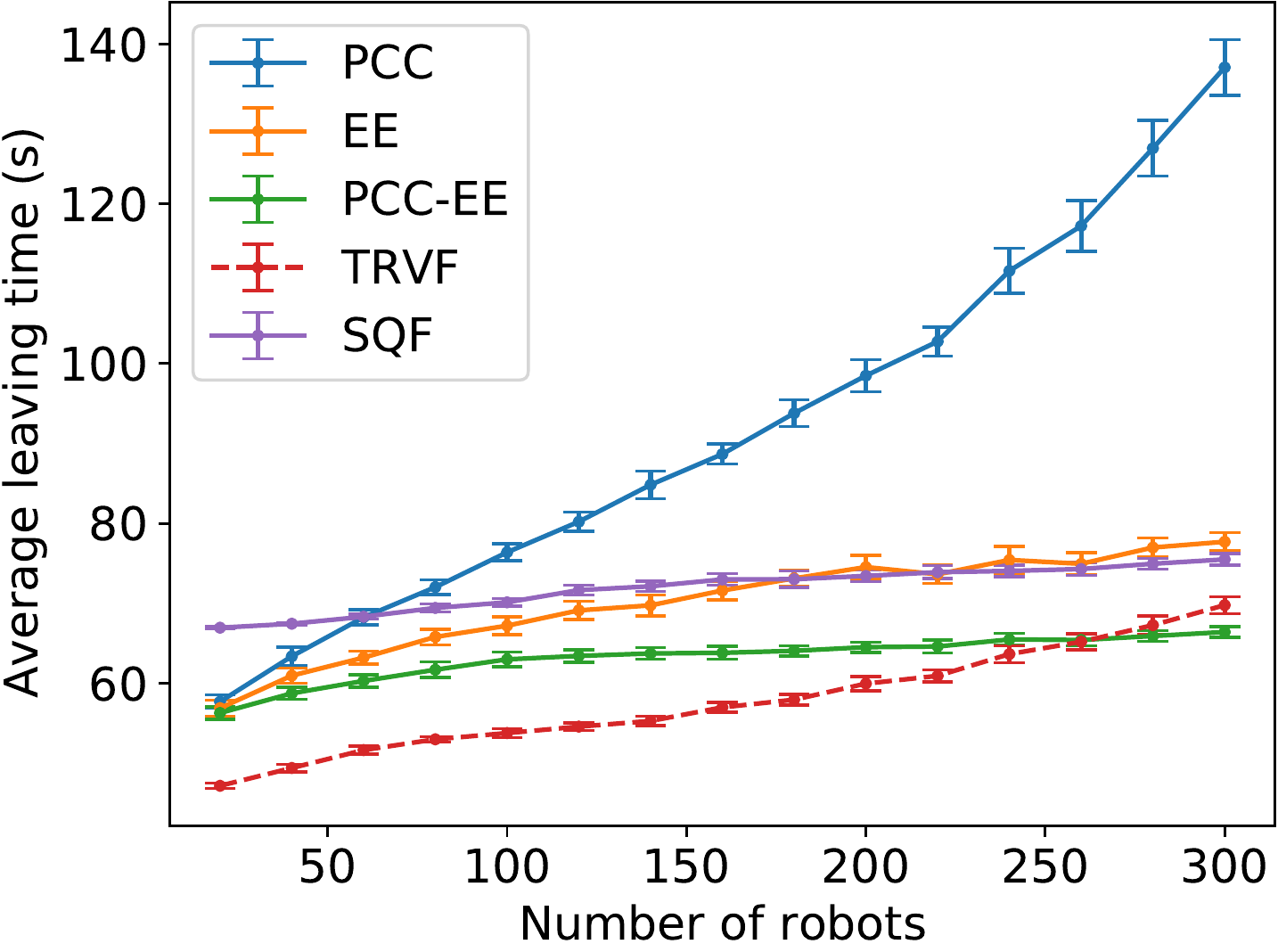}}
\subfloat[Non-holonomic]{\includegraphics[width=0.49\columnwidth]{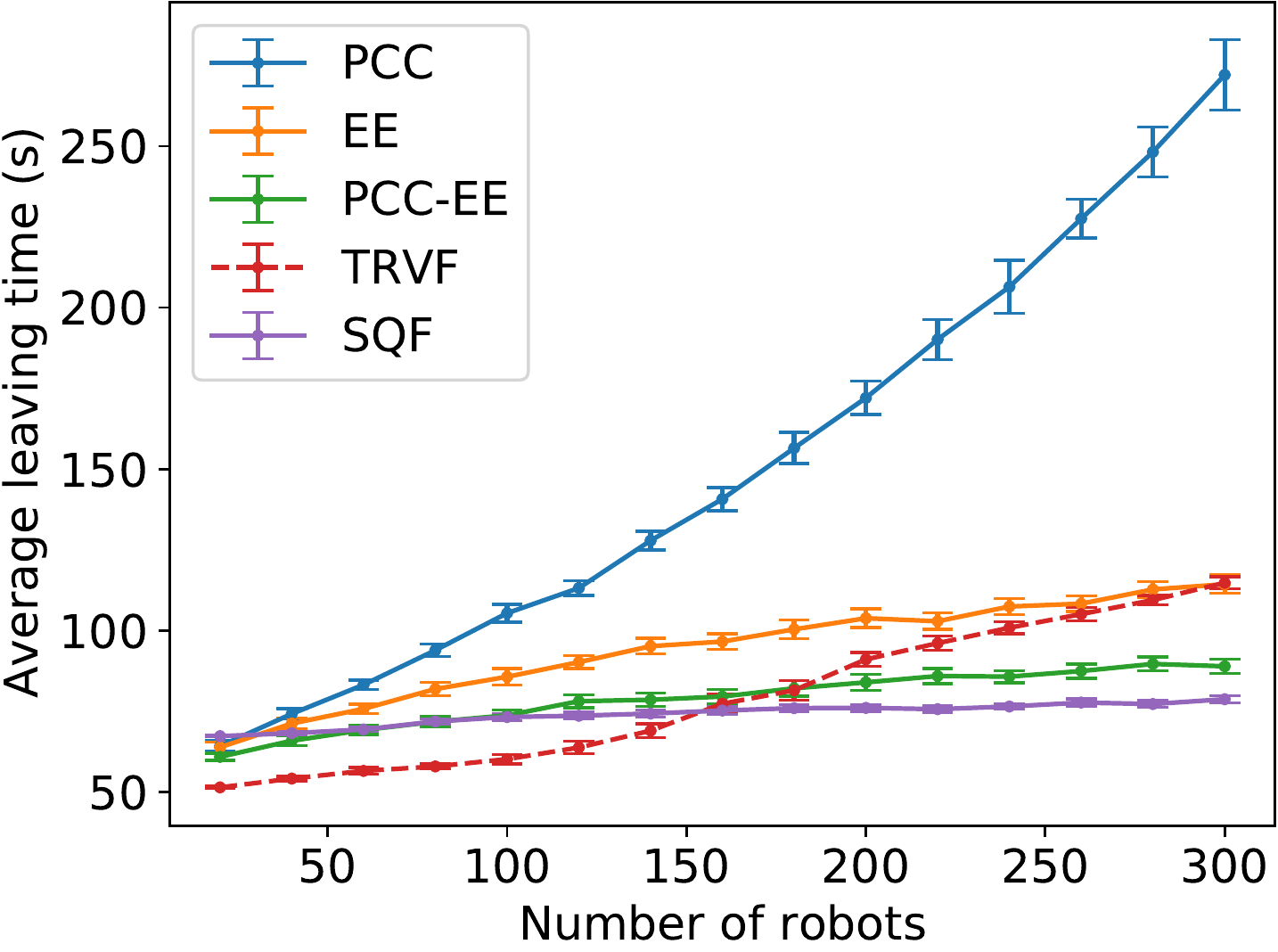}}
\caption{Comparison of the algorithms for a number of robots from 20 to 300 in steps of 20 versus the sum of the time for leaving the target area of all robots in the experiment divided by the number of robots.}
\label{fig:algsLeaving}
\end{figure}

Figure \ref{fig:algsLeaving} displays the comparison for the number of robots versus the average leaving time. In Figure \ref{fig:algsLeaving} (a), the TRVF algorithm is significantly better than all the others until 240 individuals for holonomic robots and 140 for non-holonomic ones. For holonomic robots, from 280 individuals, PCC-EE has less average leaving time than the other, while for non-holonomic robots, SQF is better from 180 individuals with statistical significance. 

As the number of robots increases, the TRVF algorithm has higher average leaving time because more robots gather next to the target area, trying to go away. For non-holonomic robots, this number is lower as they demand more time to avoid other robots when moving under non-holonomic constraints.

For holonomic robots, robots using the SQF algorithm take more time to leave the target due to the curved path that they must follow caused by the rotational force field to leave the target area. By contrast, the robots with PCC-EE follow almost a straight line from the target to leave that area, as it frees regions for leaving (as seen in Figure \ref{fig:stuck-screenshots} (b)). However, in the non-holonomic case, the robots need to make more turns or reduce linear speed to avoid others next to the target until they reach one of those regions where they can move in straight lines. With SQF, the rotational speed variation for leaving is low, and the robots can maintain linear speed most of the time, as few robots are moving in the opposite direction. 

\paragraph{Comparison of the total simulation time}

Figure \ref{fig:algsTotalTime} shows the comparison for the total simulation time. Also, note that the average time to leave the target is less than the time to reach it -- less than 15\% -- justifying the low difference in the shape of the graphs for the total simulation time and the reaching time. Apart from the shifting in the values on these graphs, SQF is better, regarding the simulation time, only starting from 280 holonomic robots (the same for the reaching time). Unpaired t-tests with $\rho = 0.01$ returned that the simulation time intervals of SQF and EE algorithms for 240 and 260 holonomic robots -- Figure \ref{fig:algsTotalTime} (a) -- have the same mean. For non-holonomic robots, the total time of SQF is less than the obtained from EE for 120 robots, as occurred for the reaching time. Likewise, unpaired t-tests showed that the simulation time intervals for 100 non-holonomic robots of SQF and EE algorithms -- Figure \ref{fig:algsTotalTime} (b) -- have the same mean (as happened for reaching time).

\subsubsection{Comparison for varying target sizes}
\label{sec:exp:varyingradius}

We now analyse the throughput and total simulation time of a varying target size using 100 robots to show that the SQF algorithm outperforms the other algorithms as the target size gets smaller. Only total simulation time is shown here because the leaving time is small compared to the reaching time, as we showed before. In this experiment, we did not use the TRVF algorithm because the restriction in the lower and upper values of $K$ \citep{arxivTheory} forbids the usage of the small values of target area radius for the influence radius used here.

\begin{figure}[t]
  \centering
  \subfloat[Holonomic.]{\includegraphics[width=0.49\columnwidth]{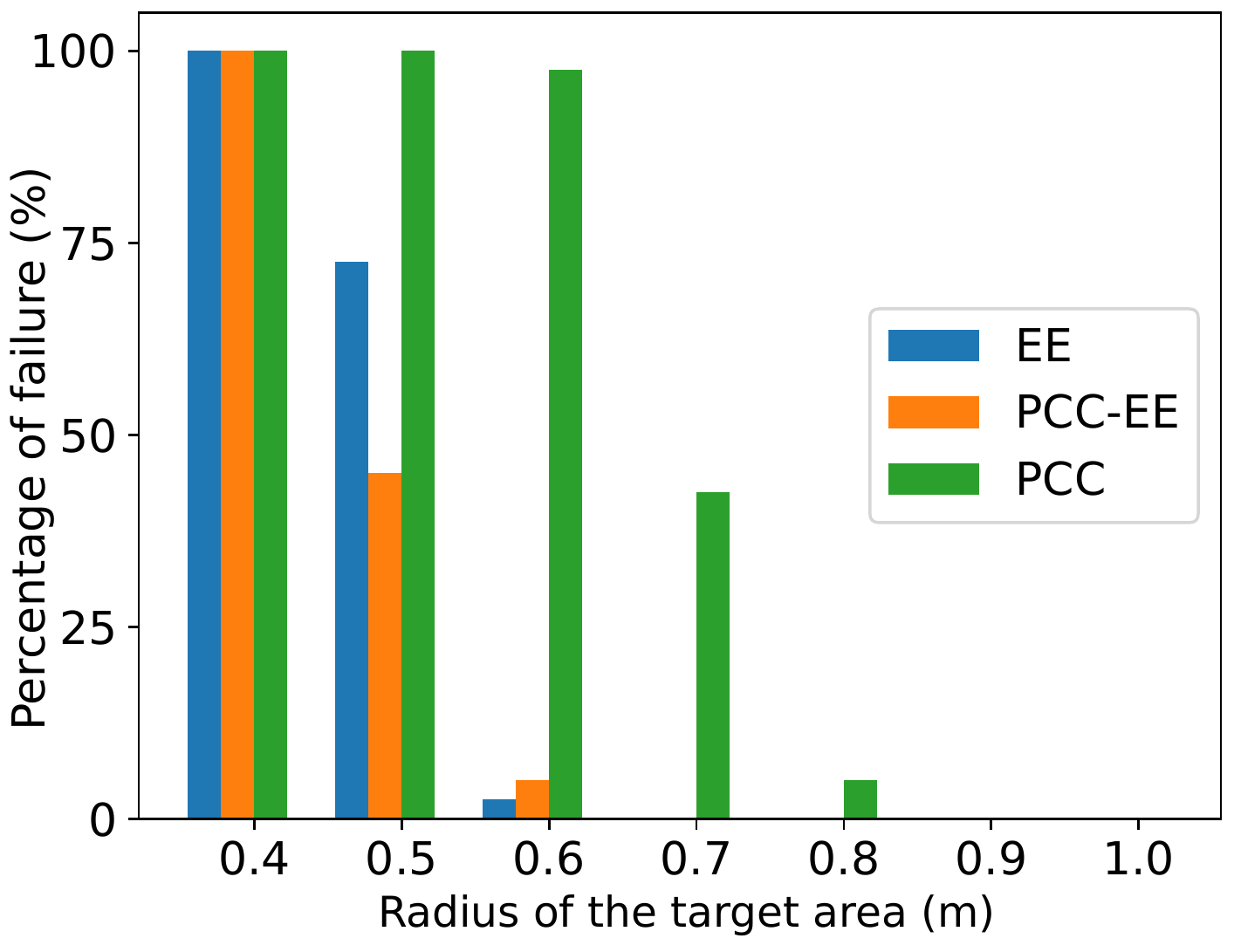}}
  \subfloat[Non-holonomic.]{\includegraphics[width=0.49\columnwidth]{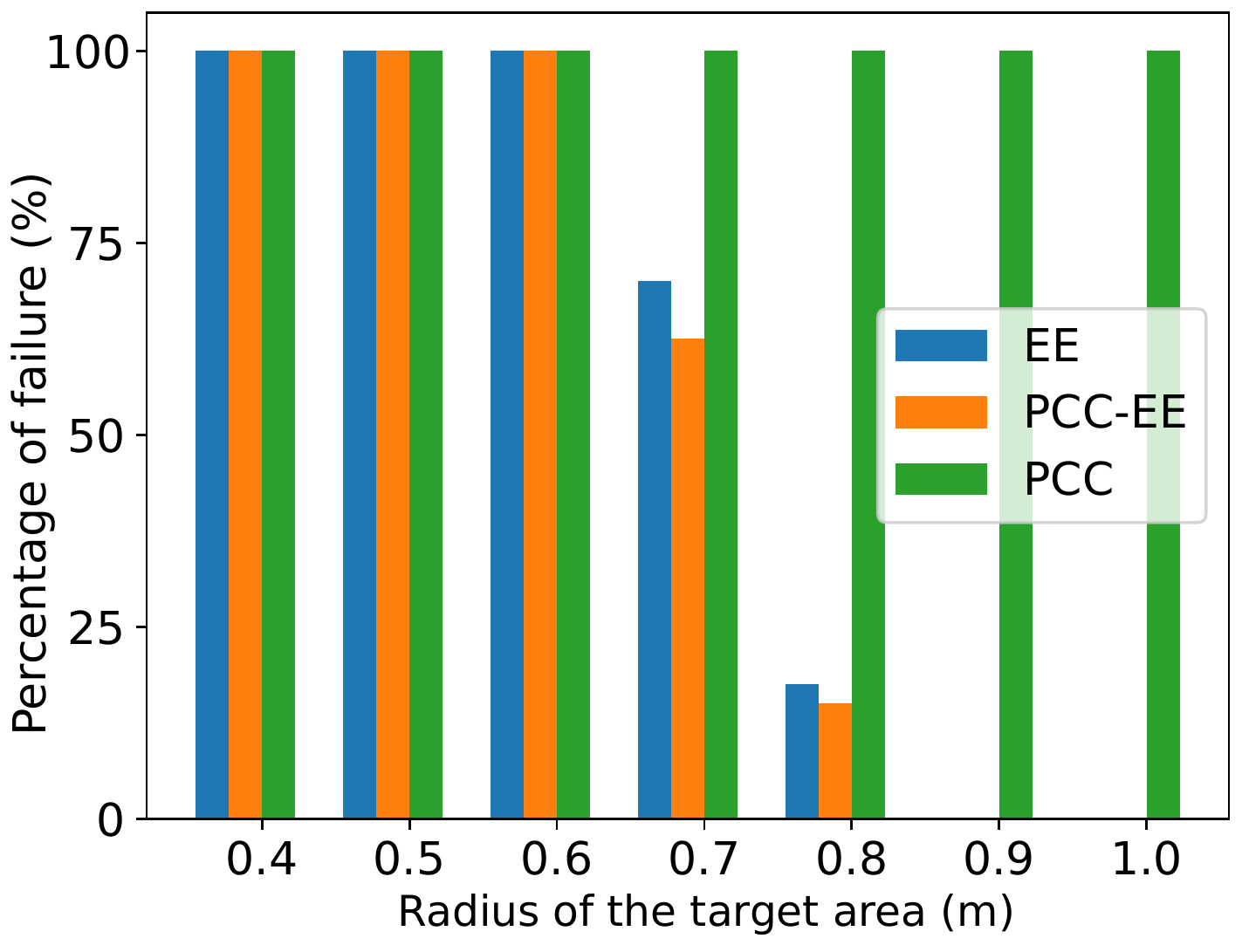}}
  \caption{The number of runs that fail to complete in a simulation time less than 20 minutes in relation to the target size and algorithm for holonomic and non-holonomic robots. SQF had zero percentage of failure in all runs.}
  \label{fig:r:target:success}
\end{figure}

First, we show that current algorithms fail to complete executions for small target sizes. For a given target area radius $s$, we respectively used $s+0.7$ m and $s+2.2$ m for the radius of free and danger regions to PCC, EE and PCC-EE algorithms \cite{Marcolino2016}. Figure \ref{fig:r:target:success} shows the percentage of failed simulations for different target sizes. For experiments with non-holonomic robots, PCC fails for all target sizes. EE and PCC-EE failed to terminate every run for a target size below $0.6$ m, but the number of unfinished experiments decreases until the radius is less than $0.9$ m. The SQF can complete all executions for the displayed target sizes.
For the holonomic robots experiments, from $s=0.9$ m all algorithms complete the execution within the time limit and the decrease in the number of failed executions per radius size is greater than the non-holonomic robots experiments.

\begin{figure}[t]
\centering
\subfloat[PCC. Available on \url{https://youtu.be/kLOLOENvnqU}.]{\includegraphics[width=0.325\linewidth]{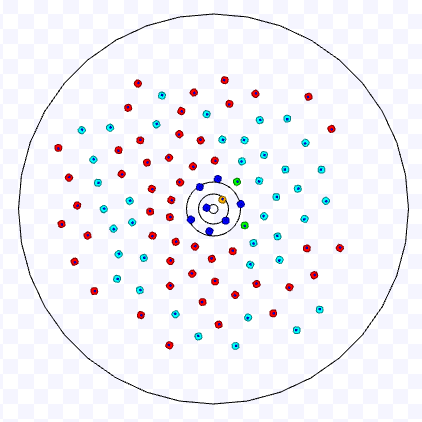}}
\hspace{1pt}
\subfloat[PCC-EE. Available on \url{https://youtu.be/UvzSqFyXB7E}.]{\includegraphics[width=0.325\linewidth]{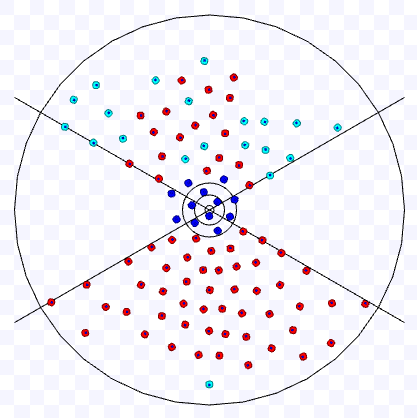}}
\hspace{1pt}
\subfloat[EE. Available on \url{https://youtu.be/BuTcsBNGCag}.]{\includegraphics[width=0.325\linewidth]{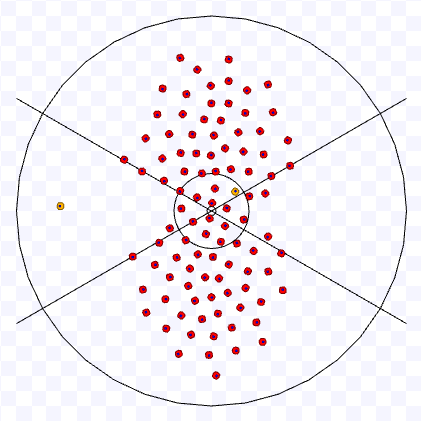}}
\caption{Executions with non-holonomic robots and $s = 0.3$ m showing situations where the robots cannot proceed within the time limit of 20 minutes. Here the colours have the same meaning as in \citep{Marcolino2016}. The robots circle around the target area for too long time. }
\label{fig:stuck-screenshots}
\end{figure}

When the radius of the target area is small, more robots tend to concentrate around it. The attractive force towards the target centre is not enough to counteract the repulsive forces from the nearby robots. This does not create a zero force vector but reduces the attraction to the target centre notably, and the robots slowly and erratically circle around the target. Figure \ref{fig:stuck-screenshots} shows examples of this situation.

\begin{figure}[t]
\centering
\subfloat[Holonomic]{\includegraphics[width=0.49\columnwidth]{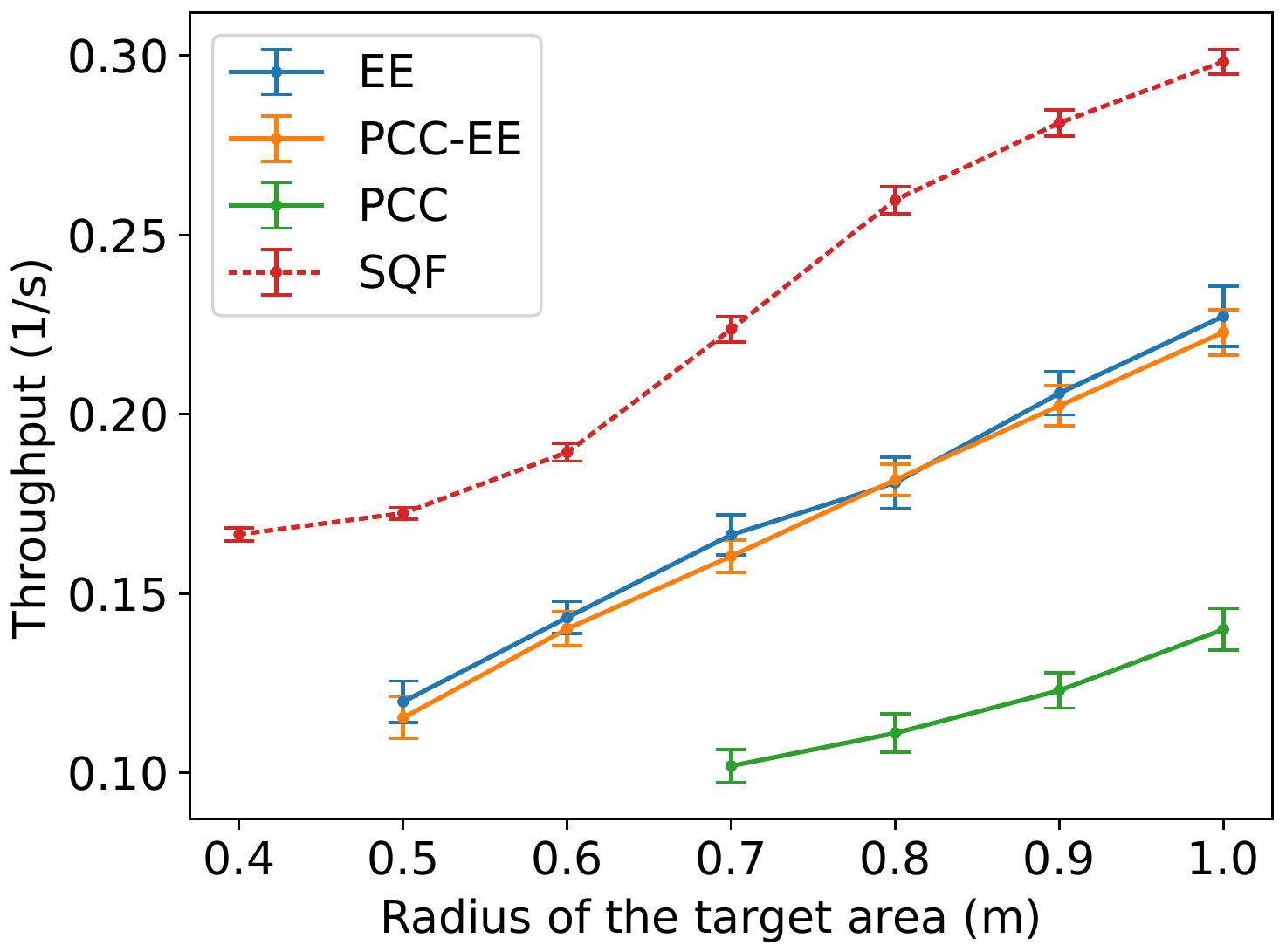}}
\subfloat[Non-holonomic]{\includegraphics[width=0.49\columnwidth]{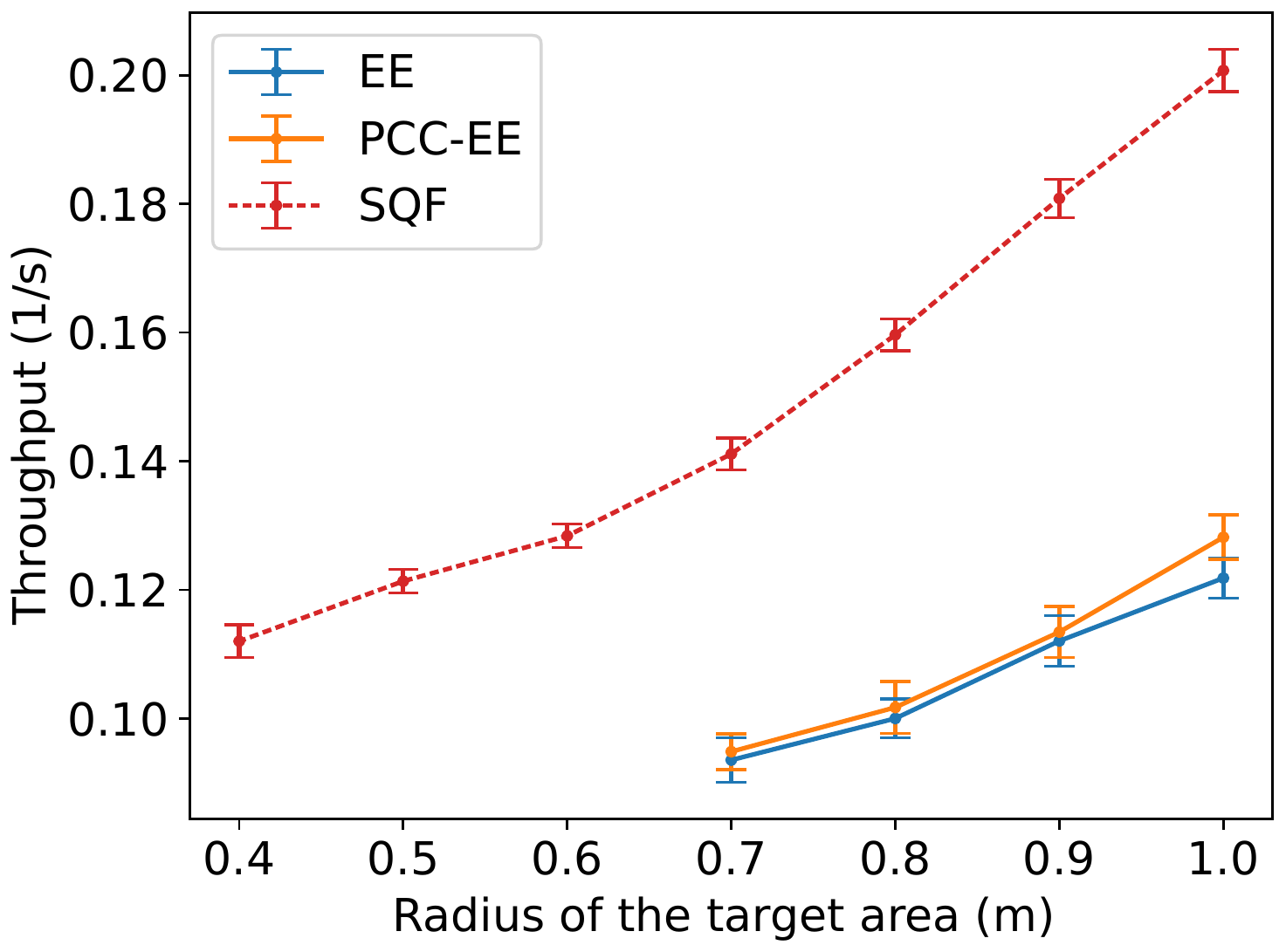}}
\caption{Throughput comparison of the algorithms for a varying circular target area radius.}
\label{fig:throughputVaryingSize}
\end{figure}

\begin{figure}[t]
\centering
\subfloat[Holonomic]{\includegraphics[width=0.49\columnwidth]{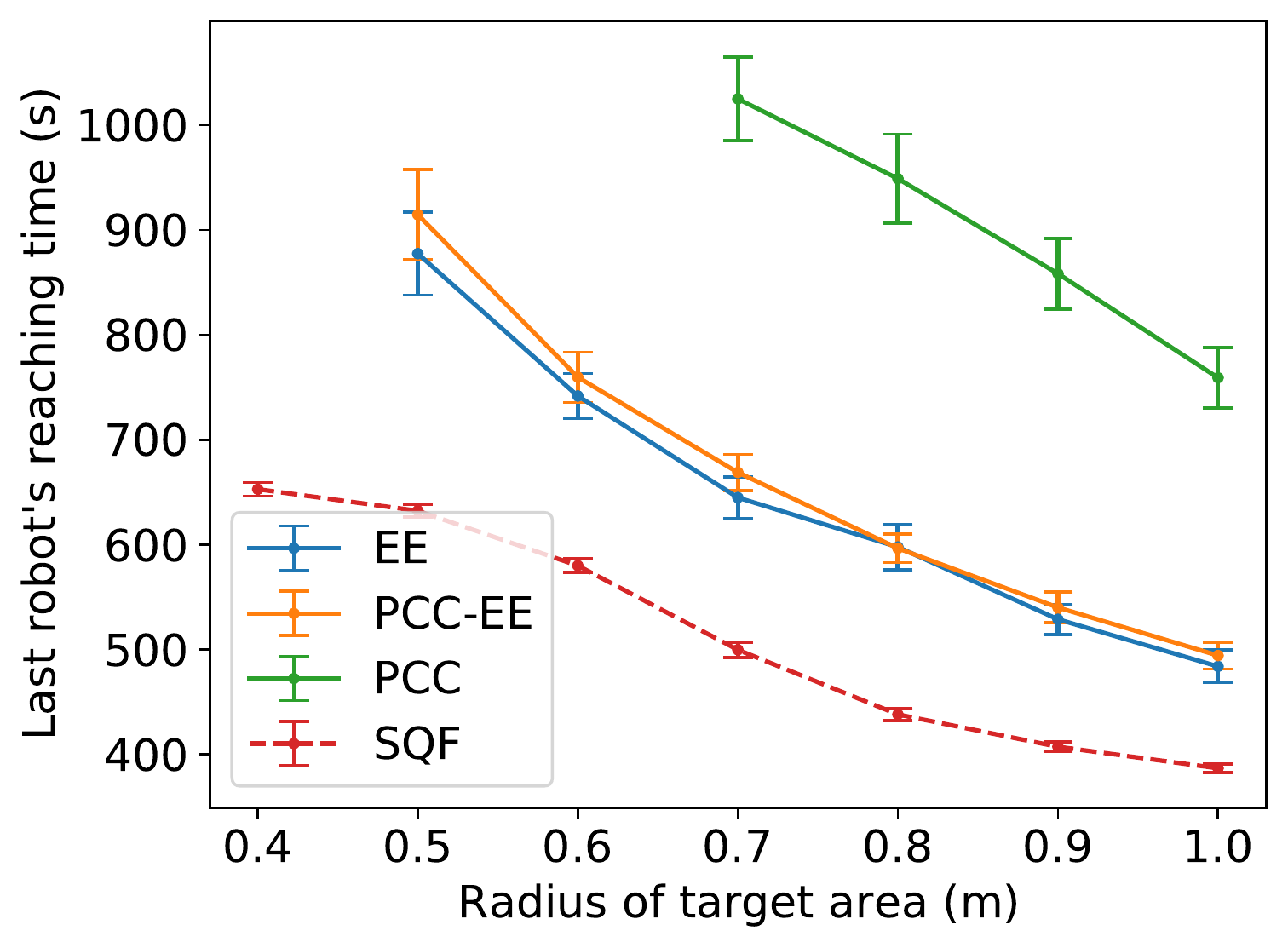}}
\subfloat[Non-holonomic]{\includegraphics[width=0.49\columnwidth]{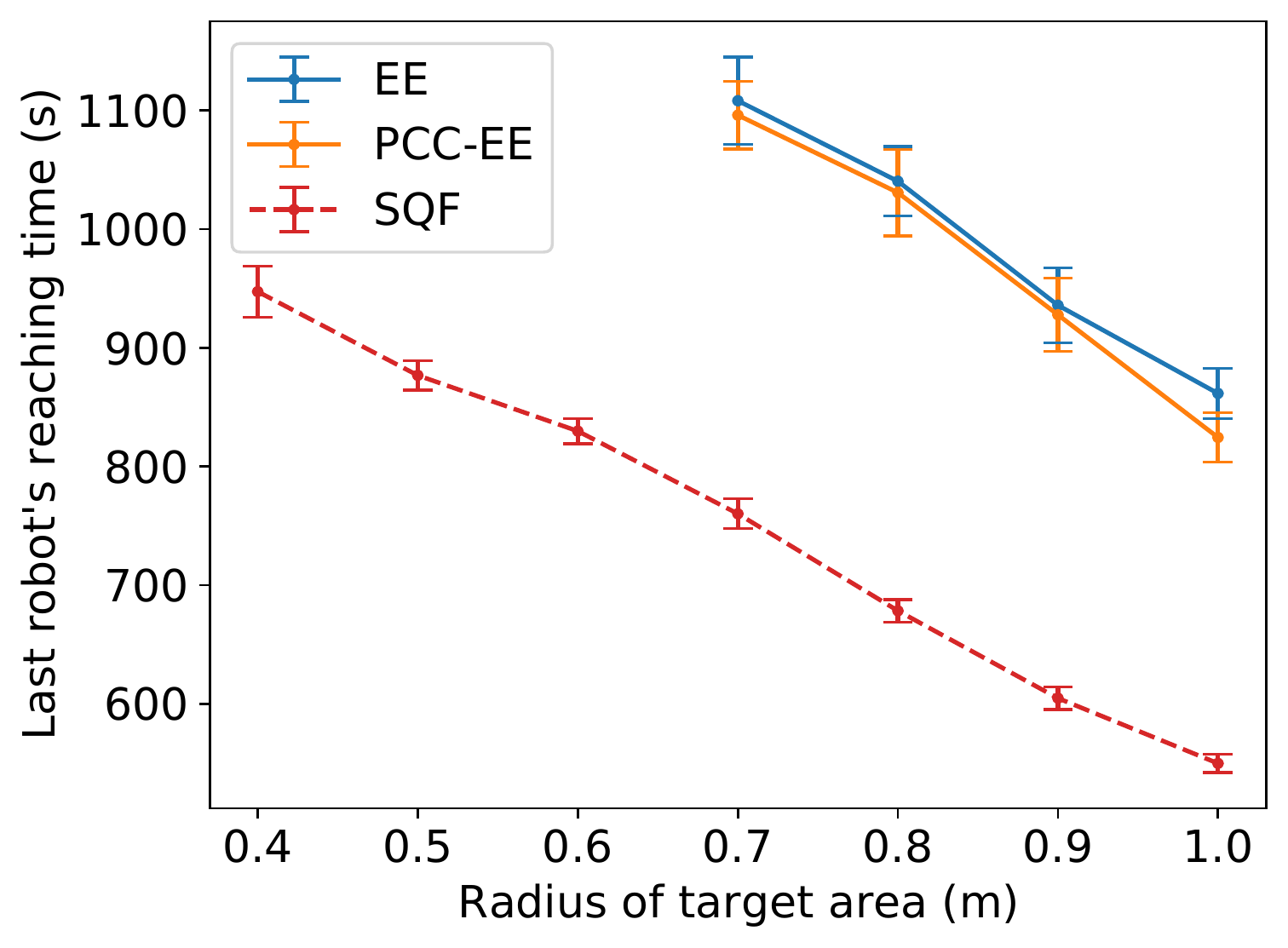}}
\caption{Simulation time comparison of the algorithms for a varying circular target area radius.}
\label{fig:timeVaryingSize}
\end{figure}

Figures \ref{fig:throughputVaryingSize} and \ref{fig:timeVaryingSize} display the throughput and time to complete by the radius of the target area. The missing points are caused by failed executions, as shown above. The confidence intervals are calculated taking in account only the successful runs. As shown in Figure \ref{fig:r:target:success} (b), PCC failed for all radius values when we used non-holonomic robots, so it is not presented in Figures  \ref{fig:throughputVaryingSize} (b) and \ref{fig:timeVaryingSize} (b). SQF significantly outperforms all other algorithms for any target size in both non-holonomic and holonomic cases.

\section{Conclusion}
\label{sec:conclusion}

The throughput of the common target area is a measure of the effectiveness of algorithms to minimise congestion in a swarm of robots trying to reach the same goal. In a concurrent work \citep{arxivTheory}, we analysed two strategies to maximise it in a common circular target region, assuming the robots running at constant maximum linear speed and having a fixed minimum distance between each other. One strategy uses a corridor with robots to enter that region. On the other, the robots follow curved trajectories to touch the region boundary. We used those strategies as inspiration to propose two new algorithms for real-world scenarios: SQF and TRVF.

We demonstrated by simulations in the Stage platform that SQF outperforms TRVF and the state-of-art algorithms PCC, PCC-EE and EE by time for reaching the target for a large number of robots, from 280 and 120 individuals for holonomic and non-holonomic robots, respectively. From these same numbers of robots, swarms running the SQF algorithm also use less time for total simulation time (i.e., reaching plus leaving the target) than the other algorithms. SQF also outperforms these algorithms concerning the average target leaving time from 180 non-holonomic robots. However, for holonomic robots, TRVF is better up to 240 individuals and PCC-EE from 280 robots. Additionally, we showed that those others might completely fail for circular target regions with an area fitting less than five times the area of a robot, and they are outperformed in throughput by SQF when they succeed.

Even though SQF has more throughput than TRVF for a number of robots greater than 140 for holonomic robots and 60 for non-holonomic ones, the comparison of their corresponding inspiration strategies (hexagonal packing and touch and run, respectively) had reversed outcomes with respect to asymptotic throughput. When we assumed constant maximum linear speed and fixed minimum distance between robots, we could provide analytical calculations of the throughput for a given time and the asymptotic throughput for the different theoretical strategies. Based solely on these calculations, we could compare which strategy is better. However, for robots using artificial potential fields, it is not straightforward to get explicit throughput equations due to the changeability of those quantities previously assumed constant. Then, in the lack of closed asymptotic equations, we performed simulations to get experimental throughput and compare algorithms for varying linear speed and inter-robot distance. As shown by the experimental data, their variation and the effect of the other robots in the trajectory affect the throughput, but the analytically calculated throughputs are still upper bounds on the ones obtained from simulation.

\section*{Acknowledgements}

Yuri Tavares dos Passos gratefully acknowledges the Faculty of Science and Technology at Lancaster University for the scholarship and the Universidade Federal do Recôncavo da Bahia for granting the leave of absence to finish his PhD.

\bibliographystyle{elsarticle-num-names}
\bibliography{swarmIntelligence}   

\end{document}